\documentclass[10pt,journal,compsoc]{IEEEtran}

\usepackage{ragged2e}
\ifCLASSOPTIONcompsoc
\usepackage[nocompress]{cite}
\else
\usepackage{cite}
\fi

\ifCLASSINFOpdf
\else
\fi
\usepackage{url} 

\usepackage{times}
\usepackage{graphicx} % more modern
\usepackage{epsfig} % less modern
\usepackage{subfigure} 

\usepackage{algorithmic}
\usepackage[vlined,ruled,linesnumbered]{algorithm2e}

\usepackage{helvet}
\usepackage{courier}

\usepackage{makecell}
\usepackage{graphicx}
\usepackage{subfigure}
\usepackage{color}
\usepackage{amsfonts}
\usepackage{amsmath}
\usepackage{amssymb}

\usepackage{multirow}
\usepackage{booktabs}

\def\etal{{\em et al.\/}\, }

\def\mF{{\mathcal F}}

\def\mL{{\mathcal L}}

\DeclareMathAlphabet\mathbfcal{OMS}{cmsy}{b}{n}
\def\0{{\bf 0}}
\def\1{{\bf 1}}

\def\bI{{\bf I}}

\def\bT{{\bf T}}
\def\bU{{\bf U}}

\def\bW{{\bf W}}

\def\bx{{\bf x}}
\def\by{{\bf y}}

\def\bx{{\bf x}}

\def\by{{\bf y}}

\def\bW{{\bf W}}

\def\ie{\mbox{\textit{i.e.}}}
\def\eg{\mbox{\textit{e.g.}}}

\usepackage{ntheorem}

\newtheorem{deftn}{Definition}

\newtheorem*{*thm}{Theorem}

\newtheorem*{*lemma}{Lemma}
\newtheorem{remark}{Remark}

\usepackage{setspace}

\usepackage{enumitem}
\def\red{\textcolor{red}}

\def\guo{\textcolor{black}}
\def\aje{\textcolor{black}}
\def\qi{\textcolor{black}}

\def\red{\textcolor{black}}

\begin{document}

% \title{The Shallow End: Multi-way Backpropagation for Training Compact Deep Networks}

\title{\huge The Shallow End: Empowering Shallower Deep-Convolutional Networks through Auxiliary Outputs}

\author{Yong~Guo,
        Jian~Chen,
        Qing~Du,
        Anton~Van~Den~Hengel,
        Qinfeng~Shi,
        Mingkui~Tan
        % Mingkui~Tan$^*$\thanks{$^*$ Corresponding author.}
        % Yongsheng~Luo
\IEEEcompsocitemizethanks{\IEEEcompsocthanksitem Yong Guo, Jian Chen, Qing Du, and Mingkui Tan are with the School of Software Engineering, South China University of Technology. E-mail: guo.yong@mail.scut.edu.cn, \{mingkuitan, ellachen, duqing\}@scut.edu.cn.
\IEEEcompsocthanksitem Anton Van Den Hengel and Qinfeng Shi are with the School of Computer Science, The University of Adelaide, Australia; and the Australian Centre for Robotic Vision. E-mail: \{anton.vandenhengel, javen.shi\}@adelaide.edu.au.
}
% \thanks{Yong Guo, Jian Chen, Qing Du, and Mingkui Tan are with the School of Software Engineering, South China University of Technology. E-mail: guo.yong@mail.scut.edu.cn, \{mingkuitan, ellachen, duqing\}@scut.edu.cn}
% \thanks{Anton Van Den Hengel and Qinfeng Shi are with the School of Computer Science, The University of Adelaide, Australia; and the Australian Centre for Robotic Vision. E-mail: \{anton.vandenhengel, javen.shi\}@adelaide.edu.au}
}

\markboth{Journal of \LaTeX\ Class Files,~Vol.~14, No.~8, September~2019}%
{Shell \MakeLowercase{\textit{et al.}}: Bare Demo of IEEEtran.cls for IEEE Journals}

	\IEEEtitleabstractindextext{%
		\begin{abstract}
		\justifying
        Depth is one of the key factors behind the success of convolutional neural networks (CNNs). Since ResNet~\cite{he2015deep}, we are able to train very deep CNNs as the gradient vanishing issue has been largely addressed by the introduction of skip connections. However, we observe that, when the depth is very large, the intermediate layers (especially shallow layers) may fail to receive sufficient supervision from the loss
        due to the severe transformation through a
        long backpropagation path. As a result, the representation power of intermediate layers can be very weak and the model becomes very redundant with limited performance. In this paper, we \guo{first} investigate the supervision vanishing issue in existing backpropagation (BP) methods. And then, we propose to address it via an effective method, called Multi-way BP \guo{(MW-BP)}, which relies on multiple auxiliary losses added to the intermediate layers of the network.
        The proposed MW-BP method can be applied to most deep architectures with slight modifications, \guo{such as ResNet and MobileNet}. \guo{Our method} often gives rise to much more compact models (denoted by ``Mw+Architecture") than existing methods. For example, MwResNet-44 with 44 layers performs better than ResNet-110 with 110 layers on CIFAR-10 and CIFAR-100.
        More critically, the resultant models even outperform 
        \guo{the light models obtained}
        by state-of-the-art model compression methods.
        Last, our method inherently produces multiple compact models with different depths at the same time, \guo{which} is helpful for model selection.
        Extensive experiments on both {image classification and face recognition} demonstrate the superiority of the proposed method.
        % Depth is one of the key factors behind the success of convolutional neural networks (CNNs). Since ResNet~\cite{he2015deep}, we can train very deep CNNs as the gradient vanishing issue has been largely addressed by the introduction of skip connections. However, when the depth is very large, the intermediate layers may fail to receive sufficient supervision due to the severe transformation through long backpropagation path. As a result, the representation power of intermediate layers can be very weak and the model becomes very redundant. In this paper, we first investigate the supervision vanishing issue in existing backpropagation (BP) methods. Then, we propose to address it via an effective method Multi-way BP, which relies on auxiliary losses added to the intermediate layers of the network. Our method often gives rise to more compact models (called “Mw+Architecture”) than existing methods. For example, MwResNet-44 with 44 layers outperform ResNet-110 with 110 layers on CIFAR-10 and CIFAR-100. More critically, the resultant models even outperform the light models obtained by state-of-the-art model compression methods. Last, our method inherently produces multiple models at the same time, which is helpful for model selection. Extensive experiments on both image classification and face recognition demonstrate the superiority of the proposed method.
		\end{abstract}
		
}

\maketitle

\IEEEdisplaynontitleabstractindextext

\IEEEpeerreviewmaketitle

\IEEEraisesectionheading{\section{Introduction}}
\IEEEPARstart{S}{ince} 2012 when AlexNet won the first place in the ImageNet competition~\cite{krizhevsky2012imagenet}, convolutional neural networks (CNNs)~\cite{lecun1989backpropagation} have been producing state-of-the-art results in many of the most challenging vision tasks including image classification~\cite{guo2018double,he2015deep,Lee2015,DBLP:journals/pami/WeiXLHNDZY16}, face recognition~\cite{schroff2015facenet,sun2015deeply,DBLP:journals/pami/RanjanPC19},
semantic segmentation~\cite{DBLP:journals/pami/ShelhamerLD17, DBLP:journals/pami/ChenPKMY18,DBLP:journals/pami/BadrinarayananK17},
and \qi{object detection}~\cite{DBLP:journals/pami/RenHG017,DBLP:journals/pami/DollarABP14}.
Moreover, deep CNNs have also become the workhorse of many other tasks and real-world applications beyond computer vision, such as natural language understanding~\cite{collobert2008unified,DBLP:journals/pami/WangWSDH18} and speech recognition~\cite{lecun2015deep}.

Recent studies~\cite{srivastava2015training,szegedy2015going} have demonstrated the importance of depth to the representation power of neural networks. Recently, the training of very deep models becomes possible (\eg, ResNet~\cite{he2015deep}), since the gradient vanishing issue has been largely addressed by introducing skip (\ie, shortcut) connections.  However, when the depth becomes large,
the model may incur training difficulties due to what we call \emph{supervision vanishing} problem.
Specifically, even with the skip connection or other advanced structures, the supervision from the loss tends to fade through a  long backpropagation path~\cite{shen2015learning}.
As a result, the intermediate layers fail to receive sufficient information from the loss, which may lead \guo{to} severe model redundancy. The existence of such redundancy \guo{often} means more parameters, larger model size, higher inference cost, more energy consumption, and/or degraded performance~\cite{ba2014deep}.
Note that in real-world applications, we have an urgent demand for efficient models with smaller model size, less energy consumption and promising performance. In this sense, how to reduce the model redundancy in CNNs while keeping/improving the performance is an important and urgent problem.

In this paper, we extensively study the supervision vanishing issue in existing BP methods and investigate why these methods would incur model redundancy even in carefully designed compact architectures. One can alleviate this issue by introducing auxiliary losses to the network~\cite{szegedy2015going}, but how to well exploit auxiliary losses in the training to obtain more compact models still remains a question. Recent studies, such as Deeply Supervised Network (DSN)~\cite{Lee2015} and GoogLeNet~\cite{szegedy2015going}, consider multiple losses as a joint loss and simply sum up the gradients from relevant losses into a joint one in \guo{backpropagation} (BP). This kind of methods  has two major limitations.
First, the multiple losses may have conflicts with each other due to their different positions in the network and simply summing them up may incur severe training difficulties. To alleviate this, one should carefully adjust the weights of losses during the training~\cite{Lee2015}, which, however, may limit its applicability to general cases. Second, these methods may still suffer from supervision vanishing and hence obtain only marginal improvement in the performance.

 %(See comparison in Section~\ref{sec:bpcomparison})

To address the supervision vanishing issue and thus reduce internal redundancy of deep models, we propose a Multi-way BP (MW-BP) method, in which we let multiple losses share one forward propagation, but conduct multiple separate backpropagations (one for each loss separately). In this way, it helps to alleviate the vanishing of supervision and obtain more compact models.
Note that in this paper we do not attempt to design compact models~\cite{howard2017mobilenets,zhang2018shufflenet} or search for some compact architectures~\cite{tan2019mnasnet,howard2019searching}.
\guo{Instead, we focus on improving the training of CNNs to obtain compact models.}
% apply our training method on various architectures to obtain more compact models by addressing the supervision vanishing issue.  by reducing internal redundancy
In fact, the proposed training paradigm can be applied to \guo{various architectures}, including both large models like ResNet~\cite{he2015deep} and \guo{lightweight} models like MobileNet~\cite{sandler2018mobilenetv2}.

%DenseNets~\cite{huang2016densely}, and Inception networks~\cite{DBLP:journals/corr/SzegedyIV16}

In the paper, we make the following contributions.
\begin{itemize}[leftmargin=*]

\item We investigate the supervision vanishing issue when training deep models using existing BP methods.  To address the issue, we exploit multiple auxiliary losses to provide additional supervision and propose an adaptive weighting scheme to alleviate the conflicts among \guo{multiple} losses.

\item
We propose a simple but effective Multi-way BP (MW-BP) method to train deep models with multiple losses.
During the training, we apply one shared forward propagation for all the losses but sequentially perform a backpropagation for each loss. In this way, the intermediate layers can receive sufficient information from each loss and hence their representation power can be significantly improved.
% which helps to obtain more compact models.
Our MW-BP can be applied to various architectures, such as ResNet~\cite{he2015deep}, DenseNet~\cite{huang2016densely},  Inception network~\cite{DBLP:journals/corr/SzegedyIV16} and MobileNet~\cite{sandler2018mobilenetv2}. We demonstrate the superiority of the proposed method with various architectures on both \textbf{image classification} and \textbf{face recognition} tasks.

%to update the model parameters using the gradients propagated from
% with various architectures, such as ResNet~\cite{he2015deep}, DenseNets~\cite{huang2016densely},  Inception networks~\cite{DBLP:journals/corr/SzegedyIV16} and MobileNet~\cite{sandler2018mobilenetv2}.

\item
The proposed method can \guo{effectively reduce the internal model redundancy and often} gives rise to more compact models {than the models trained by existing BP methods}, \ie, with fewer parameters but better performance.
For example, MwResNet-44 of \textbf{44} layers outperforms ResNet-110 of \textbf{110} layers on several benchmark data sets.
\guo{More critically, the models obtained by MW-BP even outperform the carefully compressed models obtained by state-of-the-art compression methods, in terms of both accuracy and model compactness (See Section~\ref{sec:compression}).}

	% We demonstrate the superiority of our method with various architectures on both \textbf{image classification} and \textbf{face recognition} tasks.
	
\item
Equipped with MW-BP, we inherently produce multiple models of different depths at the same time.  Surprisingly, these intermediate models often outperform their full-depth counterparts or even deeper ones trained by existing BP methods. \guo{In fact}, we can choose an appropriate one as the final model. In this sense, the proposed method is helpful for model selection.

% \guo{More critically, the compact shallow models can even outperform the pruned models by state-of-the-art compression methods, in terms of accuracy and model compactness (See Section~\ref{sec:compression}).}

\end{itemize}

\section{Related Work}\label{sec:related_Studies}

\vspace{5 pt}
\textbf{Deep Models with Multiple Losses.}
Employing auxiliary classifiers to aid in \guo{the} training has been investigated in many state-of-the-art methods.
% In DeepID~\cite{sun2015deeply}, the supervisory signals are connected to each convolution layer.
In GoogLeNet~\cite{szegedy2015going}, two auxiliary classifiers
are connected to the intermediate layers with very small weights for them to ensure the convergence (\ie, 0.3 for the auxiliary losses).
% to address the problem of vanishing gradients.
In DSN~\cite{Lee2015}, each convolution layer is associated with a classifier.
% To ensure the convergence,
To avoid the training difficulty,
DSN keeps the losses for a number of epochs and discard all but the final loss to finish the rest epochs. Unlike these methods, in the proposed MW-BP method, we do not need to \guo{set such a small weight to auxiliary losses or} discard any loss \guo{during the training}, which helps to simultaneously produce multiple models \guo{with promising performance}, with the ensuing benefits for model selection. 

\vspace{5 pt}
\noindent \textbf{Backpropagation Methods.}
Besides the standard BP method for handling a single loss, several BP variants have been proposed for dealing with multiple losses, including Joint BP~\cite{Lee2015,szegedy2015going,teerapittayanon2016branchynet} and Relay BP~\cite{shen2015learning}.
Joint BP, that has been widely used in GoogLeNet~\cite{szegedy2015going}, DSN~\cite{Lee2015} and BranchyNet~\cite{teerapittayanon2016branchynet},  essentially considers a weighted sum of multiple losses as a joint one and updates the model parameters with the joint gradients.
Another variant, called Relay BP~\cite{shen2015learning}, discards the gradients from those losses with long backpropagation paths to better preserve the supervision signal.
In both Joint BP and Relay BP, the multiple losses work jointly for the training and the gradients w.r.t. different losses are summed up in a single backpropagation.
However, even with auxiliary losses, the supervision vanishing issue can still \aje{occur} for these methods.
Unlike Joint BP and Relay BP,  in \cite{drucker1992improving}, a double backpropagation method was proposed.
Different from these methods, our MW-BP conducts a backpropagation for each loss separately.
In this way, the intermediate layers can receive sufficient supervision from the nearest losses and the supervision vanishing issue can be alleviated.

%However, both the supervision and the information of such regularization term may vanish after long backpropagation.,  where the second BP introduces a regularization term to push the input gradient from the first BP to zero

\vspace{5 pt}
\noindent \textbf{Compact Model Design.}
Recently, many attempts have been made to design compact models, \guo{such as} ResNeXt~\cite{xie2017aggregated}, MobileNet~\cite{howard2017mobilenets}, ShuffleNet~\cite{zhang2018shufflenet},
% DARTS~\cite{liu2018darts},
\emph{etc}.
Relying on ResNet~\cite{he2015deep}, ResNeXt~\cite{xie2017aggregated}  introduces group convolutions into the architecture to 
\guo{improve the model compactness.}
% obtain more compact models.
\guo{With the focus on mobile devices},
% Focuses on mobile devices,
% \guo{To apply deep models on mobile devices,}
MobileNet~\cite{howard2017mobilenets} employs depthwise separable convolution \guo{to build lightweight networks.}
ShuffleNet~\cite{zhang2018shufflenet} uses a channel shuffle operation to reduce the model size and inference complexity.
\guo{Instead of designing compact architectures}, we focus on devising an effective training method to obtain more compact models.
Empirically, the proposed MW-BP method exhibits good compatibility with various architectures and can 
\guo{produce more compact models than the ones trained by existing BP methods.}
% further enhance the performance by improving the compactness of models.

\vspace{3 pt}
\noindent \textbf{Model Compression Methods.}
Recently, many efforts have been made to obtain compact models via model compression techniques.
For example, one can prune unimportant channels based on a pretrained CNN \guo{and introduce sparsity into the filters of convolution}
% such as ThiNet, DCP, etc
\cite{luo2017thinet,zhuang2018discrimination,he2017channel,li2016pruning,he2019filter}.
Li \etal utilize an $\ell_1$-norm criterion to prune unimportant filters~\cite{li2016pruning}.
In~\cite{he2019filter}, He \etal propose to use geometric median of the filters to perform channel pruning.
% In~\cite{li2016pruning}, Li \etal measured the importance of channels using the sum of absolute value of weights.
Unlike these methods, we seek to develop an effective training algorithm to produce compact models.
% we seek to train a shallow but compact model, which has much fewer parameters but with comparable or even better performance than the deeper ones, without any further compression.
{More critically, the resultant models trained by MW-BP even outperform the carefully compressed models obtained by state-of-the-art model compression methods} (See results and comparisons in Section~\ref{sec:compression}).

%Interestingly, our proposed method is able to obtain compact models with promising performance without any further compression techniques (See results in Section~\ref{sec:compression}).

\section{Supervision Vanishing in Deep Networks}\label{sec:supervision_vanishing}
In this section, we  study the issue of supervision vanishing
{in the training of deep networks}.

% in training deep CNNs.
Without loss of generality, we consider an $L$-layers network \aje{that}
conducts \emph{forward propagation} for any layer $l$ by
%\begin{equation}\label{eq:residual}
%{\bx_{l + 1}} = h(\lambda_l\bx_{l} + {\mF_l}({\bx_l},{{\bW}_{l}})),
%\end{equation}
\begin{equation}\label{eq:residual}
\by_l = \lambda_l \bx_l + {\mF_l}({\bx_l},{{\bW}_{l}}), ~~~\bx_{l+1}=h(\by_l),
\end{equation}
where $\lambda_l \in \{0,1\}$. Here, $\bx_l$ and $\bx_{l+1}$ denote the input and output of the $l$-th layer, respectively;
$\by_l$ denotes the intermediate feature before activation;
$h$ is a nonlinear activation function (\eg, Rectified Linear Unit (ReLU)~\cite{nair2010rectified} or Sigmoid function); and $\mF_l$ denotes a transformation function (\eg, convolution operation) parameterized by $\bW_l$. \guo{When $\lambda_l=0$, Eqn. (\ref{eq:residual}) represents the forward propagation process of plain deep networks, such as AlexNet~\cite{krizhevsky2012imagenet} and VGG~\cite{simonyan2014very}.}
When $\lambda_l=1$, there is a shortcut connection between the $l$-th and $(l+1)$-th layer. The shortcut connection, an effective technique to avoid the gradient vanishing issue in BP, enables us to train very deep models that are known as the residual networks~\cite{he2015deep}.
% Moreover,  the network is known as the residual network if $\lambda_l=1, \forall l \in \{1, ..., L\}$~\cite{he2015deep}.

In practice,
% one can train deep networks by applying stochastic gradient descent (SGD)~\cite{Wilson2003The} to update the parameters $\{\bW_l\}$.
one can use stochastic gradient descent (SGD)~\cite{Wilson2003The} to update the parameters $\{\bW_l\}_{l=0}^{L-1}$.
Let $\xi$ be the loss function, {the gradient of $\xi$ w.r.t. $\bW_l$ can be computed by}
\begin{equation}\label{eq:gradientw}
    \frac{\partial \xi}{\partial \bW_l} = \frac{\partial \xi}{\partial \bx_{l+1}} \frac{\partial \bx_{l+1}}{\partial \bW_l},
\end{equation}
\guo{where $\frac{\partial \xi}{\partial \bx_{l+1}}$ denotes the gradient propagated from $\xi$ to some intermediate layer}.
By applying the chain rule according to Eqn.~(\ref{eq:residual}),
% the gradient of $\xi$ w.r.t. $\bx_l$
such gradient for any layer $l$  can be written as
\begin{equation}\label{eq:backpropagation}
\begin{aligned}
\frac{\partial \xi} {\partial \bx_l}
&=
\frac{\partial \xi} {\partial \bx_L} \left( \frac{\partial \bx_L} {\partial \bx_{L-1}}
\cdots \frac{\partial \bx_{l+1}} {\partial \bx_l} \right) \\
&=
\frac{\partial \xi} {\partial \bx_L}
\prod\limits_{j = l}^{L - 1} \frac{\partial \bx_{j+1}}{\partial \by_j} \frac{\partial \by_j}{\partial \bx_j}  \\
&= \frac{\partial \xi} {\partial \bx_L}
\prod\limits_{j = l}^{L - 1} \bT_j(\bW_j),
\end{aligned}
\end{equation}
where
\begin{eqnarray}\label{eq:transform}
\bT_j(\bW_j) = \frac{\partial \bx_{j+1}}{\partial \by_j} \Big( { \lambda_j \bI + \partial_{\bx_j}  {\mF_j({\bx_{j}},\bW_{j})}} \Big).
\end{eqnarray}

%Based on $\frac{\partial \xi} {\partial \bx_{l+1}}$

\begin{deftn}[\textbf{Supervision Information}]
	We define $\frac{\partial \xi}  {\partial \bx_l}$,  the partial gradient of $\xi$ w.r.t. $\bx_l$,  as the \emph{supervision information} obtained from the loss. From Eqn.~(\ref{eq:backpropagation}), the partial gradient $\frac{\partial \xi} {\partial \bx_l}$ contains two parts, namely  $\frac{\partial \xi} {\partial \bx_L}$ and  $\prod_{j = l}^{L - 1} \bT_j(\bW_j)$,
	where the term $\frac{\partial \xi} {\partial \bx_L}$ is directly related to the loss $\xi$.
    % \guo{The first term $\frac{\partial \xi} {\partial \bx_L}$ is the supervision information directly obtained from the loss $\xi$, and the second term $\prod_{j = l}^{L - 1} \bT_j(\bW_j)$ is a series of transformations along the backpropagation path.}
\end{deftn}

Note that each $\bT_j(\bW_j)$ is a transformation matrix that transforms $\frac{\partial \xi} {\partial \bx_L}$ a bit. Then, the term $\prod_{j = l}^{L - 1} \bT_j(\bW_j)$ will transform the gradient $\frac{\partial \xi} {\partial \bx_L}$ through a series of layers from the final layer to the $l$-th layer. When $(L{-}l)$ is large, the transformation $\prod_{j = l}^{L - 1} \bT_j(\bW_j)$ can be too severe and
%\guo{make the shallow layers hard to receive sufficient supervision from $\frac{\partial \xi} {\partial \bx_L}$.}
make the component $\frac{\partial \xi} {\partial \bx_L}$ negligible in $\frac{\partial \xi}  {\partial \bx_l}$. 
In this case, the shallow layers cannot receive sufficient supervision from the final loss $\xi$, since the gradient $\frac{\partial \xi}  {\partial \bx_l}$ has very limited information from $\xi$ due to the severe transformation of the long-path backpropagation. We call this phenomenon the \textbf{supervision vanishing issue}. As a result, the intermediate layers (especially the shallow layers) may have \guo{limited representation power}, which will incur severe \textbf{internal redundancy} in deep models.

% transformation 造成信息损失，导致监督信息vanish

\section{Multi-way Backpropagation for Deep Models with Auxiliary Losses} \label{sec:backpropagation}

% to help intermediate and shallow layers obtain additional supervision.

\begin{figure}[tp]
	\begin{center}
		\includegraphics[width=1\columnwidth]{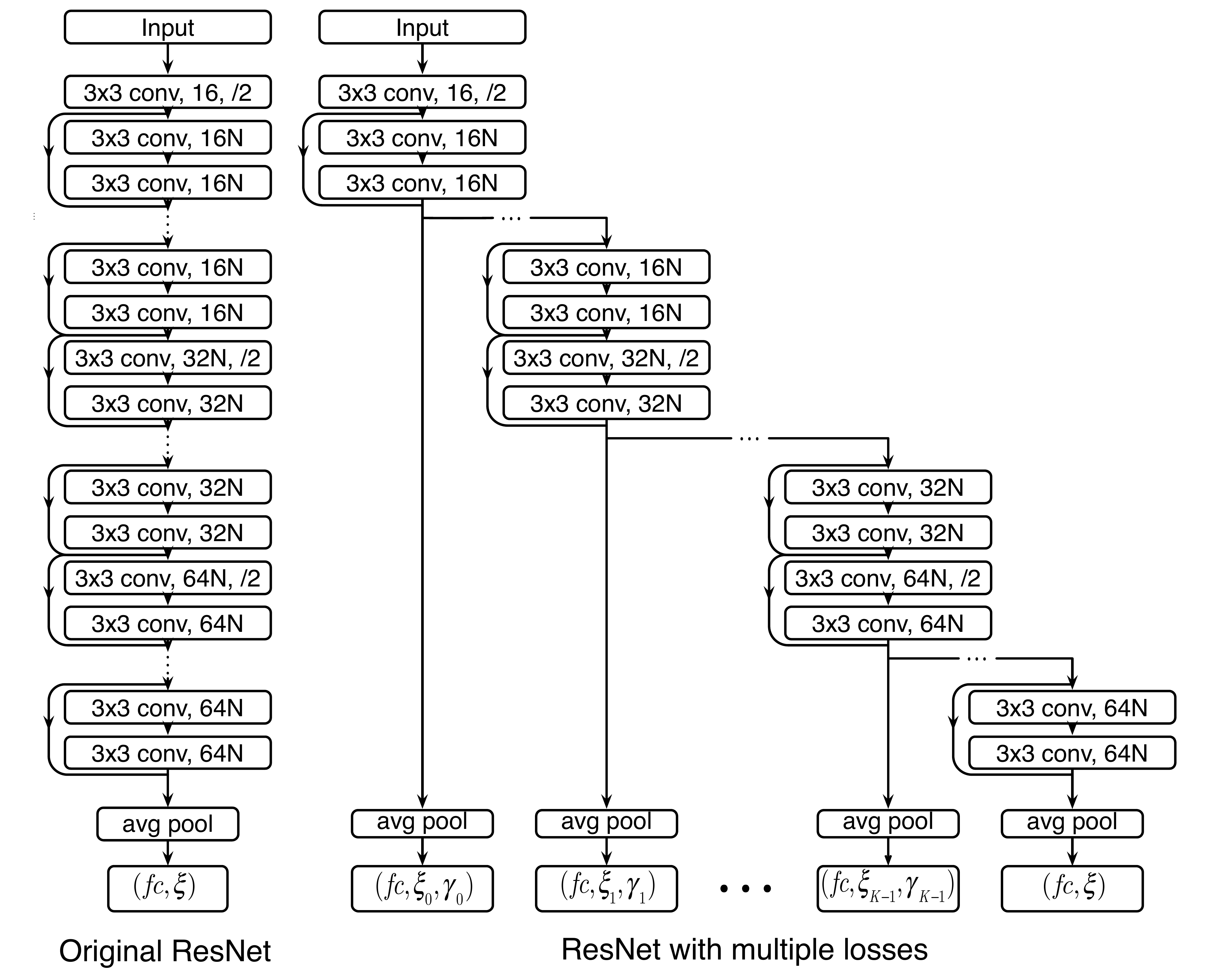}
	\end{center}
	\caption{Architecture with auxiliary losses. Taking ResNet for example, we put $K$ additional losses $\{\xi_i\}_{i=0}^{K-1}$ evenly to the network, with each loss being built on top of an average pooling layer.  Here, the $K$  losses are weighted by $\{\gamma_i\}_{i=0}^{K-1}$, $fc$ indicates the fully connected layer,  $\xi$ (also called $\xi_{K}$) denotes the final loss  and $N$ denotes the width of networks.}
	\label{fig:ausresnet}
\end{figure}

\subsection{Deep Model with Auxiliary Losses}\label{sec:model_structure}
%\guo{Since each loss is always associated with a fully connected layer that produces the prediction output, we also call the module with an auxiliary loss at intermediate layers an auxiliary output.}

As mentioned in Section~\ref{sec:supervision_vanishing}, the \guo{standard} BP with a single loss may incur supervision vanishing issue and lead to redundant models. To address this, it is natural to introduce auxiliary losses to the network to provide additional supervision for  shallow layers, similar to DSN~\cite{Lee2015} and GoogLeNet~\cite{szegedy2015going}. However, how to avoid the possible conflicts among different losses and well exploit the information from auxiliary losses to train compact models are still open questions.

%To address the supervision vanishing issue, one may introduce multiple auxiliary losses to the network
%\guo{to provide additional supervision for the intermediate and shallow layers}.

Taking an $L$-layer ResNet for example, as shown in Fig.~\ref{fig:ausresnet},
we introduce $K$ auxiliary losses to the network, with  each being built on the top of an average pooling layer.
Including the final loss $\xi$, we have $\widehat{K} {\small=} K{\small +}1$ losses in total.
% \guo{we introduce $K$ auxiliary losses to the network and thus have $(K+1)$ losses in total.}
We can either apply the same form of the final loss $\xi_K$ to each auxiliary loss $\{\xi_i\}_{i=0}^{K-1}$ or exploit other forms of losses for them.
% From Fig.~\ref{fig:ausresnet},
\guo{For convenience, we use $L_i$ to indicate the layer to which the $i$-th loss is connected.}
Note that each loss $\xi_i$ is associated with a model of depth $L_i$. Thus, with the multiple losses, we can inherently obtain multiple models of different depths, as shown in Fig.~\ref{fig:ausresnet}.

\subsubsection{Adaptive Weighting Scheme for Auxiliary Losses}\label{sec:weighting}

The auxiliary losses, however, may have conflicts even with the same form, which may incur training difficulties or inferior performance.
Specifically, the gradients from different losses may have different directions. Moreover, since the model with fewer layers has less representation power, the shallow-layer losses can be very large even after many iterations of training, which may lead to training difficulties. \guo{To alleviate this}, we develop an adaptive weighting scheme for different losses.

% \guo{By default, we set the weight for the final loss to $\gamma_K {\small =} 1$.}
Since the auxiliary losses are not equally important, we should impose different confidence, denoted by $\{\gamma_i\}_{i=0}^{K-1}$, over them. \guo{By default, we set $\gamma_K {\small =} 1$ for the final loss.}
% where $0 \leq \gamma_i \leq 1$.
\guo{For the auxiliary losses $\{\xi_i\}_{i=0}^{K-1}$}, in general, the losses at deeper layers should be more important, since the features at deeper layers often have better representation power. In this sense, we use $\gamma_i = (\frac{L_i}{L_{K-1}})^\nu$ to reflect such difference, where
% $L_k$ is the index of the layer to which the $k$-th loss is added and
$\nu > 0$ is the decaying rate of $\gamma_i$. In practice, we observe that if $\gamma_i<0.01$, the effect of $\xi_i$ becomes negligible. We thus use the following adaptive rule to adjust the weights for different losses:
\begin{equation}
\gamma_i = \max \left( 0.01, \left( \frac{L_i}{L_{K{-}1}} \right)^\nu \right), \forall i \in \{0, \ldots, K{-}1\}.
\label{eq:weighting}
\end{equation}

By setting $\nu > 0$, the weights of shallower losses will be smaller, which is helpful for the training convergence and thus improve the overall performance. Here, we suggest setting $\nu \in [1/2, 2]$.
In practice, similar to the adjustment of learning rate,  we may apply an adaptive strategy to adjust $\nu$ during the training process (See details in Section~\ref{exp:nu}).

% we should set $\nu$ to a small value (\eg, 1/2) at the end of training
% In practice, one may change $\nu$ dynamically during the training.
% In particular,

\subsubsection{Number of Auxiliary Outputs}
There remains a question \aje{regarding} how many auxiliary losses should be introduced.   We observe that, \red{adding too many outputs would hamper the performance due to the conflicts of losses and also significantly increase the training complexity (See discussions of $K$ in Section~\ref{exp:num_outputs})}. Without loss of generality, we can introduce an auxiliary loss every $\tau=\lceil L/ \widehat K \rceil $ layers, where $\tau\geq 5$.

%When $\widehat K = L$, each layer will be associated with an auxiliary outputs.
%\guo{In practice, we observe that adding up to five losses is sufficient to achieve good performance.} , {where $L$ denotes the depth of the network and $\widehat K$ denotes the number of losses}

\subsection{Existing BP Methods for Multiple Losses}\label{sec:existing_bp}
Several BP methods have been proposed to train networks with auxiliary losses,  \red{\eg,} Joint BP~\cite{Lee2015,szegedy2015going,teerapittayanon2016branchynet} and Relay BP~\cite{shen2015learning}.

% In both methods, all losses work jointly for the training and the gradients w.r.t. different losses are summed up in a single backpropagation.
% In this sense, we call such BP scheme Single-way BP.
%\guo{Moreover, we also construct two simple variants based on Multi-way BP, called Na\"{i}ve Multi-way BP and Inverse Multi-way BP, to better investigate the proposed method.}
%Let $\partial_{\bx_l} \mL$ be the gradient induced by all losses w.r.t. $\bx_l$. Let ${\xi}_k$ be the loss of the $k$-th output and ${\xi_{K+1}}$ be the loss of the final output.
% of the joint loss ${\cal{L}}$ w.r.t. $\bx_l$  can be computed by
\subsubsection{Joint BP}
Joint BP  considers minimizing a joint objective function of multiple losses~\cite{Lee2015,szegedy2015going}:
\begin{equation} \label{eq:companion_obj}
{\mL} = \sum\limits_{i=0}^{K} \gamma_i \xi_i .
\end{equation}
% Upon the chain rule, \guo{given some index $k \in \{0, \ldots, K\}$}, for any shallow layer $l$ (where $L_{k-1} < l < L_{k}$, with $L_{-1}=-\infty$),
% the gradient of $\mL$ w.r.t. $\bx_l$ becomes
% % can be computed by
% Upon the chain rule,
\guo{With the focus on the $k$-th loss},
% for any shallow layer $l$ (where $L_{k-1} < l \leq L_{k}$),
the gradient of $\mL$ w.r.t. $\bx_l$ (where $L_{k-1} < l \leq L_{k}$)
can be computed by
\begin{equation}
\begin{aligned}
\red{\frac{\partial \mL}{\partial \bx_l} = \sum\limits_{i=k}^{K} \gamma_i \frac{\partial \xi_i} {\partial {\bx_{L_i}} } \prod\limits_{j = l}^{L_i - 1} \bT_j (\bW_j).}
\end{aligned}
\label{eq:jointgradient}
\end{equation}
From Eqn.~(\ref{eq:jointgradient}), Joint BP considers the information from all \guo{the} losses by summing up the gradients. However, it has several limitations. First, the deep-layer losses  (often
with large weights) may dominate the gradients in Eqn.~(\ref{eq:jointgradient}) and  the gradients from shallow-layer losses (often with very small weights) can be negligible. \guo{Thus}, similar to the \guo{standard} BP, the transformation $\prod_{j = l}^{L_i - 1} \bT_j(\bW_j)$ for the deep-layer losses may cause information vanishing at intermediate layers~\cite{shen2015learning}, resulting in significant information loss in Eqn.~(\ref{eq:jointgradient}).

Second, due to the possible conflicts among losses, the gradients w.r.t. $\bx_l$ from different losses may have different directions. As a result, the gradient in Eqn.~(\ref{eq:jointgradient}) can be inaccurate, which may incur training difficulties. To alleviate this, one should carefully adjust the weights $\gamma_i$ for the auxiliary losses.
For example, in DSN~\cite{Lee2015},  the weights for auxiliary losses gradually decrease to zero during the training. However, decreasing the weights of auxiliary losses fails to fully exploit the auxiliary losses and thus hampers the overall performance (See results in Section~\ref{sec:bpcomparison}).

\begin{figure*}[t]
    \centering
   	\subfigure[Multi-way Backpropagation]{\label{fig:multiBP}
		\includegraphics[width = 0.97\columnwidth]{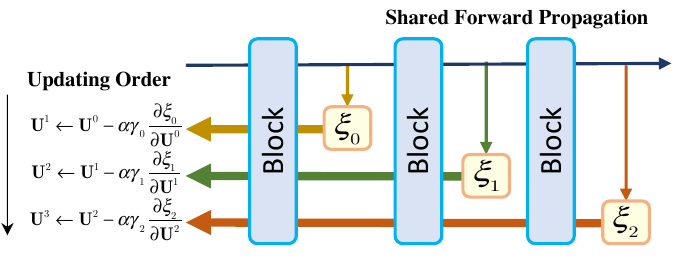}
	}
\hspace{0.05in}
	\subfigure[Joint Backpropagation]{\label{fig:singleBP}
   		\includegraphics[width = 0.97\columnwidth]{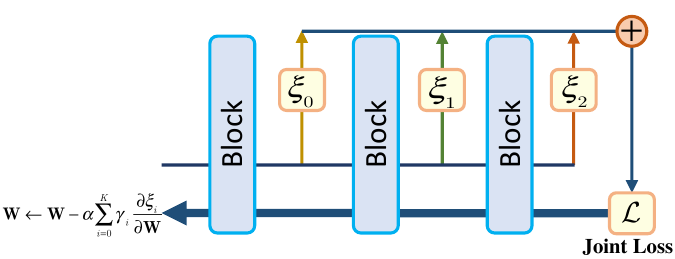}
   	}
	\caption{A simple demonstration of  Multi-way Backpropagation (\textit{left}) and Joint Backpropagation (\textit{right}) for training deep networks with auxiliary losses.
% 	\guo{Taking the backpropagation w.r.t. $\xi_1$ for example, the red box denotes the layers to be updated but kept unchanged in the backpropagation w.r.t. $\xi_0$.}
	}\label{fig:BP}
\end{figure*}

%which will further incur information loss in Eqn.~(\ref{eq:jointgradient}).
%In fact, in GoogLeNet~\cite{szegedy2015going}, it sets a large weight to the last loss and small weights to the auxiliary losses. , especially when they are with large weights

\subsubsection{Relay BP}
%(caused by long backpropagation path of deep-layer losses)
%
To alleviate the possible information loss issue  in Joint BP,
Shen \etal proposed a
% new variant called
Relay BP~\cite{shen2015learning} method
that discards the gradients propagated from deep-layer losses far away from a considered layer. \guo{With the focus on the $k$-th loss},
% for any layer $l$ (where $L_{k-1} < l \leq L_k$),
the gradient w.r.t. $\bx_l$ (where $L_{k-1} < l \leq L_k$) becomes
\begin{equation}
% \partial_{\bx_l} \mL
\frac{\partial \mL}{\partial \bx_l}
= \sum\limits_{i=k}^{k+c} \gamma_i \frac{\partial \xi_i} {\partial {\bx_{L_i}} } \prod\limits_{j = l}^{L_i - 1} \bT_j (\bW_j),
\label{eq:relaygradient}
\end{equation}
which means the $l$-th layer only receive the gradients from $\{\xi_i\}_{i=k}^{k+c}$, where $c \geq 1$ is a constant. If $c \geq K-k$, Relay BP is reduced to Joint BP. When $c < K-k$,  those gradients with long paths are discarded.
However, without considering deep-layer losses, the shallow layers will tend to over-fit the nearby losses, which may deteriorate the representation power of the whole network \guo{(See results in Section~\ref{sec:bpcomparison})}.

% by discarding the gradients propagated from those losses far away from the layer Thus, Relay BP can alleviate the information loss issue incurred by long propagation path.

\begin{algorithm}[t]
	\centering
	\caption{MW-BP for training deep networks.}
	\label{alg:multiupdate-algorithm}
	\begin{algorithmic}[1]\small
		\REQUIRE  Model parameters: $\{\bW_l\}_{l=0}^{L-1}=\{\bW_0,\ldots,\bW_{L{-}1}\}$;\\
		~~~~~~~Output positions: $\{L_i\}_{i=0}^{K} = \{L_0,\ldots,L_{K}\}$;\\
		~~~~~~~Weights for different losses $\{\gamma_i\}_{i=0}^{K} = \{\gamma_0, \ldots, \gamma_{K}\}$;\\
		~~~~~~~Input data: $\bx_0$; Learning rate: $\alpha$.
		\STATE ~\//\// \textbf{Shared Forward Propagation} for multiple outputs
		\FOR{$l=0$ to $L{-}1$}
		\STATE Compute $\bx_{l+1} = h\left({\lambda_l \bx_{l}} + \mF_l({\bx_{l}},{{\bW}_{l}})\right)$;
		\IF{$l \in \{L_i\}_{i=0}^{K}$}
		\STATE Compute the loss $\xi_i$; \\
		\ENDIF
		\ENDFOR
		\STATE ~\//\// \textbf{Multi-way Backward Propagation} \\
% 		\FOR{$l=0$ to $L$}
% 		\STATE $\bU^0_l = \bW_l$ \\
% 		\ENDFOR
		\STATE Let $\bU^0_l \leftarrow \bW_l$, $\forall l \in \{0, \ldots, L{-}1\} $;\\
% 		$\{\bU^0_l\}$ = $\{\bW_l\}$ \\
		\FOR{$i=0$ to $K$}
% 		\STATE Let $\bU^i_l \leftarrow \bW_l$, $\forall l \in \{0, \ldots, L_i\} $;\\
		\STATE Conduct backpropagation w.r.t. $\xi_i$ to compute $\{\frac{\partial \xi_i} {\partial \bU_l^{i}}\}_{l=0}^{L_i}$;
% 		\STATE Propagate gradient from $\xi_i$ with the \textbf{latest} $\{\bU_l^{i-1}\}$; \\
		\STATE Update model parameters that are relevant to $\xi_i$ by \\
		~~~~~~~~$\bU_l^{i+1} {\leftarrow} \bU_l^{i} {-} \alpha \gamma_i \frac{\partial \xi_i} {\partial \bU_l^{i}}$, $\forall l \in \{0, {\ldots}, L_i\} $;
		\STATE Update model parameters that are irrelevant to $\xi_i$ by \\
		~~~~~~~~~~$\bU_l^{i+1} {\leftarrow} \bU_l^{i}$, $\forall l \in \{L_i{+}1, {\ldots}, L{-}1\} $;\\
%		\FOR{$l = 0$ to $L_i$}
%		\STATE Update the parameters by $\bU_l^{i+1} \leftarrow \bU_l^{i} - \alpha \cdot \gamma_i \frac{\partial \xi_i} {\partial \bU_l^{i}} $; \\
%		\ENDFOR
		\ENDFOR\\
% 		\FOR{$l=0$ to $L$}
% 		\STATE $\bW_l = \bU^{K+1}_l$ \\
% 		\ENDFOR
        \STATE Let $\bW_l \leftarrow \bU^{K{+}1}_l$, $\forall l \in \{0, \ldots, L{-}1\} $;\\
        % $\{\bW_l\} = \{\bU^{K+1}_l\}$ \\
	\end{algorithmic}
\end{algorithm}

% To address the supervision vanishing issue and avoid the conflicts among losses in existing BP methods,
% \guo{
% we propose an effective training method called Multi-way BP.}
% We show the training details in Algorithm~\ref{alg:multiupdate-algorithm}. For clarity, we also show the comparison between Multi-way BP and Joint BP in Fig.~\ref{fig:BP}.

\subsection{Multi-way Backpropagation}\label{sec:multiwaybp}

Using auxiliary losses is helpful for providing additional supervision information for intermediate layers. However, due to the possible conflicts among  multiple losses, simply summing up the gradients propagated from multiple losses into a joint one (as done by Joint BP and Relay BP) may not achieve promising performance, as the conflicts are inherently ignored. Moreover, the supervision vanishing issue may still happen in Joint BP due to the long propagation path.

To address the above issues and well exploit the information from all the losses, we propose a simple yet effective method, called Multi-way BP \guo{(MW-BP)}, to train deep neural networks with multiple losses. The overall scheme  is shown in Algorithm~\ref{alg:multiupdate-algorithm}, which consists of \textbf{one shared forward propagation} and \textbf{multiple backward propagations} in each iteration. Note that, when performing the multiple backpropagations, we update the model parameters but keep the features and losses unchanged.  As will be explained, this paradigm can  effectively alleviate the possible conflicts among different losses \guo{and hence can address the supervision vanishing issue}.

In Algorithm~\ref{alg:multiupdate-algorithm}, similar to existing BP methods, in each iteration, we conduct a forward propagation to update the features and compute all the losses in $\{\xi_i\}_{i=0}^{K}$. However, when updating the model parameters, unlike Joint BP and Relay BP, we conduct multiple backpropagations (one for each loss in $\{\xi_i\}_{i=0}^{K}$) in a sequential manner.
\guo{An illustrative comparison between Joint BP and MW-BP can be found in Fig.~\ref{fig:BP}.}

% as shown in Algorithm~\ref{alg:multiupdate-algorithm}.
%Actually, we update the features and compute all the losses within one  forward propagation process.
%starting from the shallowest loss $\xi_0$,

Taking the $i$-th loss $\xi_i$ for example, we compute the gradient of the $l$-th layer ($\forall l\leq L_i$) via the backpropagation and update the model parameters via batch stochastic gradient descent (SGD).
Let $\bx_l$ be the input feature of the $l$-th layer and $\bU_l^i$ be the updated parameters
after the ($i$-1)-th backpropagation,
with $\bU_l^0$ being initialized by $\bW_l$.
For the $i$-th loss $\xi_i$,
\guo{we seek to update the parameters $\{\bU_l^i\}_{l=0}^{L_{i}}$ that include both the updated parameters $\{\bU_l^i\}_{l=0}^{L_{i-1}}$ by the $(i{-}1)$-th loss and the unchanged parameters of the layers between $L_{i-1}$ and $L_{i}$.}
% (\eg, the red box in Fig.~\ref{fig:BP} when $i{=}1$).
\guo{Based on $\{\bU_l^i\}_{l=0}^{L_{i}}$,}
we update the model parameters
% $\{\bU_l^{i+1}\}_{l=0}^{L_i}$
by
\begin{equation}\label{eq:update_U}
    \bU_l^{i+1} \leftarrow \bU_l^{i} - \alpha \cdot \gamma_i \frac{\partial \xi_i} {\partial \bU_l^{i}},
\end{equation}
where  $\alpha$ denotes the learning rate and $ \frac{\partial \xi_i} {\partial \bU_l^{i}} = \frac{\partial{\xi_i}}{\partial \bx_{l+1}}  \frac{\partial\bx_{l+1}} {\partial{\bU_l^{i}}}. $
By applying the chain rule,
the gradient $\frac{\partial \xi_i}{\partial \bx_l}$ for any layer $l$
($ l {\leq} L_i$) can be computed by
\begin{equation}\label{eq:multibp_single}
\frac{\partial{\xi_i}}{\partial \bx_l}
= \frac{\partial \xi_i} {\partial {\bx_{L_i}} } \prod\limits_{j = l}^{L_i - 1} \bT_{j}(\bU_j^{i}),
\end{equation}
with $\bT_{j}(\bU_j^{i})= \frac{\partial \bx_{j+1}}{\partial \by_j} (  \lambda_j \bI + \partial_{\bx_j}  {\mF_j({\bx_{j}},{{\bU}_{j}^{i}})} ).$ 

\begin{remark}\label{remak:order}
According to Eqn. (\ref{eq:update_U}), we use $-\frac{\partial \xi_i} {\partial \bU_l^{i}}$ as the search direction, which means the $(i{+}1)$-th update is dependent on the $i$-th update. As will be explained, this is very important for MW-BP (See Section~\ref{sec:characteristic_order}). In fact, one may use $-\frac{\partial \xi_i} {\partial \bW_l}$ as the search direction for the $(i{+}1)$-th update, \ie, each backpropagation is independent of each other. This strategy, however, is essentially the Joint BP \guo{method} in Fig. \ref{fig:singleBP}.
\end{remark}

\begin{remark}\label{remak:shared}
Unlike existing BP methods in which the forward and backward propagations are often performed in pairs, in MW-BP, we apply a shared forward propagation for multiple updates, namely we do not update the features and losses after each backward propagation. As will be explained in Section~\ref{sec:characteristic_share}, this is valid and also essential in boosting the performance and reducing the training complexity. 
\end{remark}
  
In the following, we will investigate more characteristics of MW-BP.

\subsection{Characteristics of MW-BP}\label{sec:characteristics}
Relying on the training paradigm in Algorithm \ref{alg:multiupdate-algorithm}, MW-BP has several characteristics over existing methods.

First, MW-BP can effectively address the supervision vanishing \guo{issue} occurred in the standard BP. 
In the standard BP, the long backpropagation path in Eqn.~(\ref{eq:backpropagation}) w.r.t. a single loss tends to incur the supervision vanishing issue. However, for MW-BP, we consider multiple backpropagations (one for each loss) to update model parameters in Eqn.~(\ref{eq:update_U}). Clearly, from Eqn.~(\ref{eq:multibp_single}), any intermediate layer $l$ (especially the shallow layers) can receive sufficient supervision from their nearby losses (with $L_i\geq l$).

Second, MW-BP can also effectively avoid the loss conflict issue in Joint BP and Relay BP. As shown in  Eqn.~(\ref{eq:jointgradient}) and Eqn.~(\ref{eq:relaygradient}), Joint BP and Relay BP simply sum up the gradients of the related losses into a joint one. In this way, the conflicts among losses may affect the model performance (See Section~\ref{sec:existing_bp} for details). Unlike these two methods, in MW-BP, we conduct the backpropagation for each loss. Thus, the summation process is avoided. As a result, the risk of loss conflict can be greatly reduced. However, to well address the supervision vanishing \guo{issue} and \guo{the} loss conflict issue,  the importance of the \textbf{order of conducting backpropagations}  and the \textbf{shared forward propagation} should be highlighted.

% 第一顺序，第二，结合顺序，讲不会有vanishing的原因：单独更新，更新后的参数会反映shallow到deep的知识
% 第三，讲固定feature和shared forward的好处

\subsubsection{\textbf{The order of conducting backpropagations}} \label{sec:characteristic_order}
In MW-BP, we conduct  multiple backpropagations from the loss $\xi_0$ to $\xi_K$ in a sequential way. A primary reason is that, a deeper loss in general is more important than a shallower loss (See Section~\ref{sec:model_structure} for details). 
Note that shallower models often have less representation power than deeper models. Thus, the attempt to fit shallower losses may introduce errors or distortions to the whole network~\cite{Lee2015}. 
Fortunately, according to Eqn. (\ref{eq:update_U}),
the $(i{+}1)$-th BP is built on the $i$-th model update (See Remark \ref{remak:order}). 
In this way, the errors brought by the model update w.r.t. a shallow loss can be corrected by the model update w.r.t. the deeper losses, which helps to 
obtain a good whole model and promising intermediate models. 
% improve the overall performance and obtain promising intermediate models. 
In other words, the order of backpropagations is essential for addressing the loss conflict issue.

%1. 交代背景：我们MW-BP中shared forward是怎么做的
%2. 在existing BP中，forward后面都会跟着一个backward
%3. 如果我们也使用 a forward followed by a backward。当前loss更新完之后，下一个forward使用到的参数既包括了之前更新过的参数，也包括没更新过的参数（图2中的红框），模型前后不一致就会影响deep loss的计算和训练。
%4. 实验中发现这样做的话，deep loss会下降很快，但是testing performance很差 (see Fig 3)
%5. shared forward就可以保留所有loss原有的分辨能力，实验结果也更好，而且不会增加cost

%, we compute the features and losses within one forward propagation and use them in multiple backpropagation

\subsubsection{\textbf{The shared forward propagation}} \label{sec:characteristic_share}
As stated in Remark \ref{remak:shared}, the shared forward propagation is one of \guo{the} key features in MW-BP, which means that we do not update 
both features and losses after each backpropagation, even though a part of model parameters have been changed. 
In fact, upon the updating order in MW-BP, if updating the features and losses, the previous updates may highly affect the update w.r.t. deeper losses. For example, the deeper/final losses may decrease too quickly at the beginning epochs, which may incur gradient vanishing issue when updating deep layers and thus deteriorate the overall performance (See results in Section~\ref{sec:bpcomparison}).
Moreover, as previously mentioned, the model update w.r.t. a shallow loss may bring in errors  (which can be distortions to the whole network). However,
% the forward propagation (which updates both features and losses) 
{if updating the features and losses, the corresponding forward propagation}
may propagate the errors to the deeper losses and hamper the correction effect of the model update w.r.t. them. Last, by avoiding multiple forward propagations, the shared forward propagation can significantly reduce the training complexity.

\subsection{More Discussions}\label{sec:variants}
% 先总结特点
% forward 不精确
% possible varients, for example
% Naive BP， loss 降低 导致 gradient vanishing issue
% order matters
% inverse multi-way BP
% 深层到底层

To verify the above arguments and further analyze the proposed MW-BP method, in the following, we introduce several possible variants by considering the paired forward-backward propagation and/or different updating orders.

\begin{figure*}[t]
	\centering
	\subfigure[Testing error of different models.]{
		\includegraphics[width = 0.63\columnwidth]{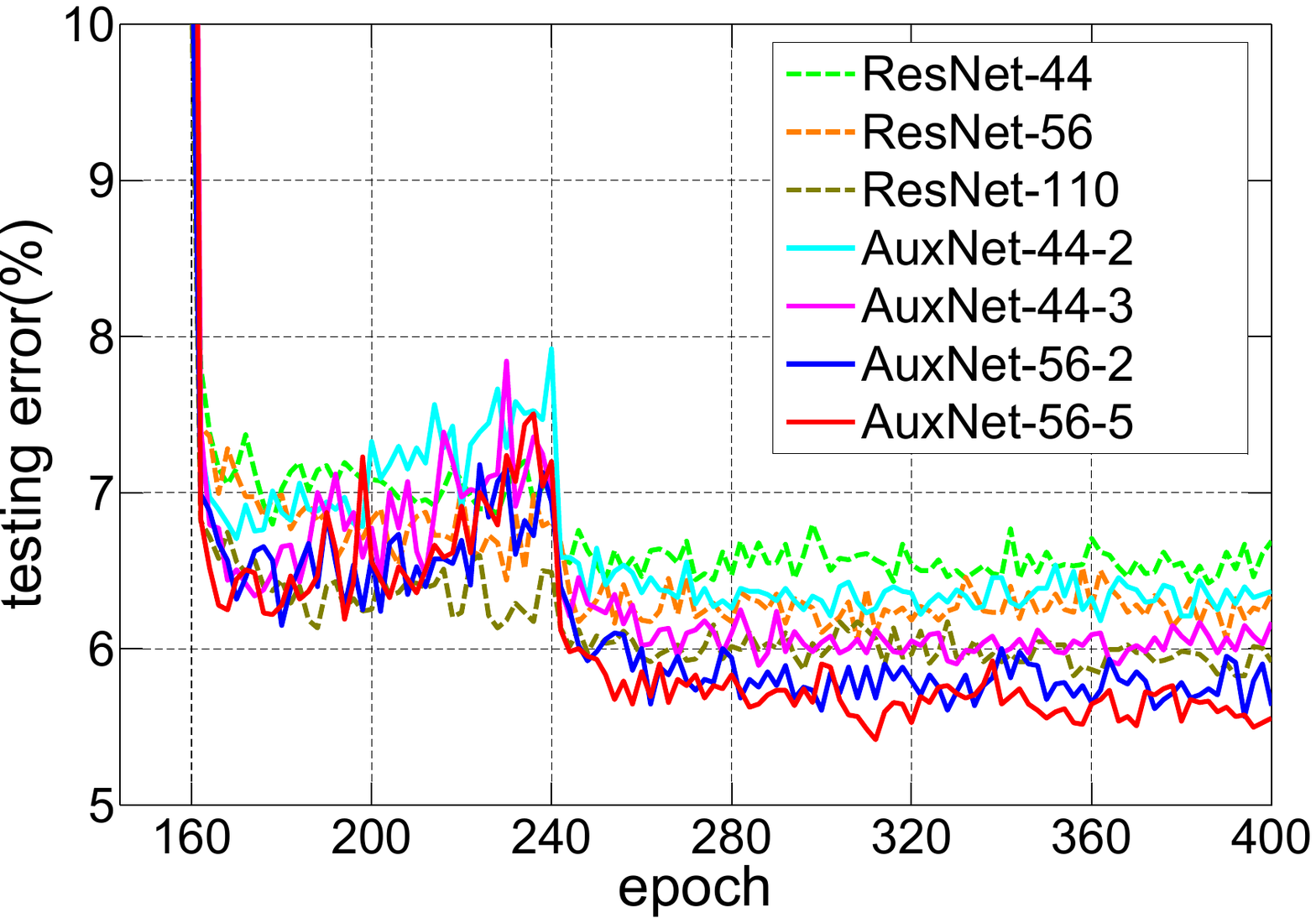}
		\label{fig:cifar10test}
	}
	\subfigure[Testing error of different BP methods.]{
		\includegraphics[width = 0.63\columnwidth]{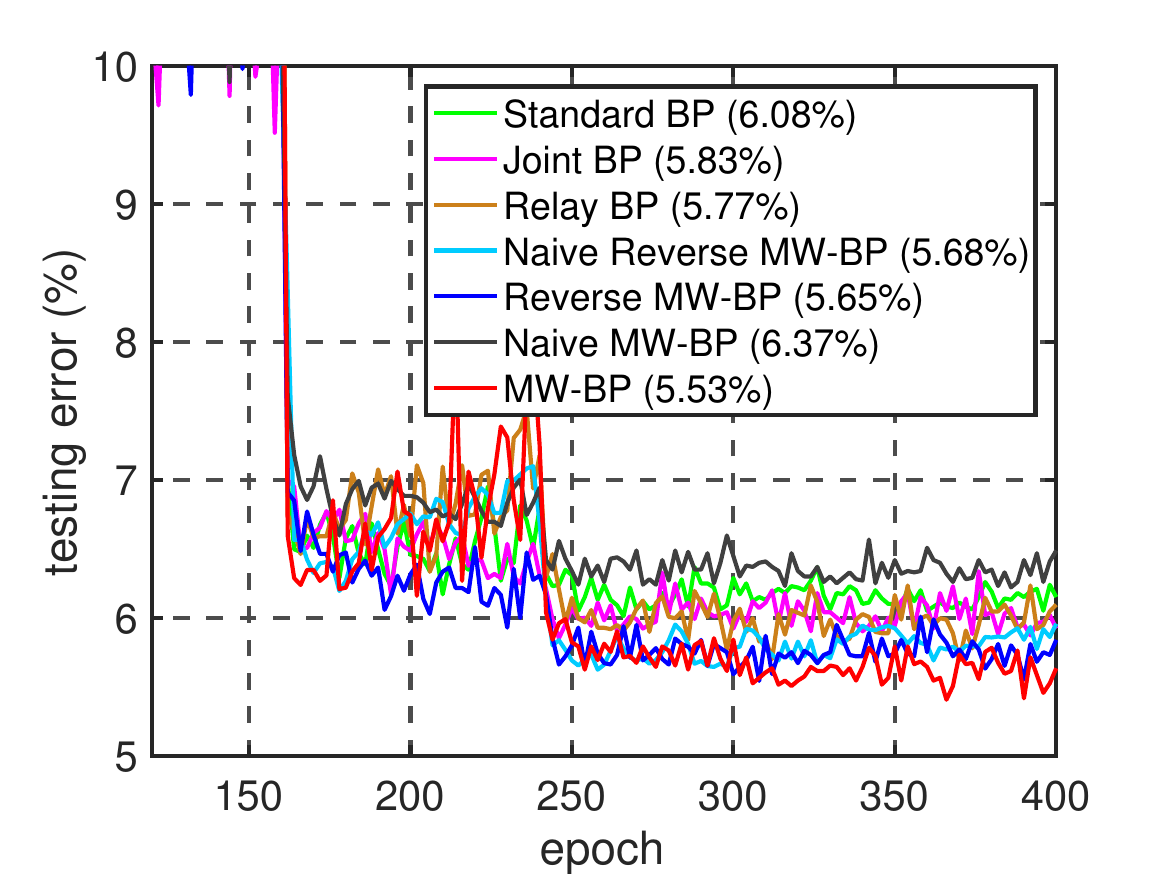}
		\label{fig:bptest}
	}
	\subfigure[Training error of different BP methods.]{
		\includegraphics[width = 0.63\columnwidth]{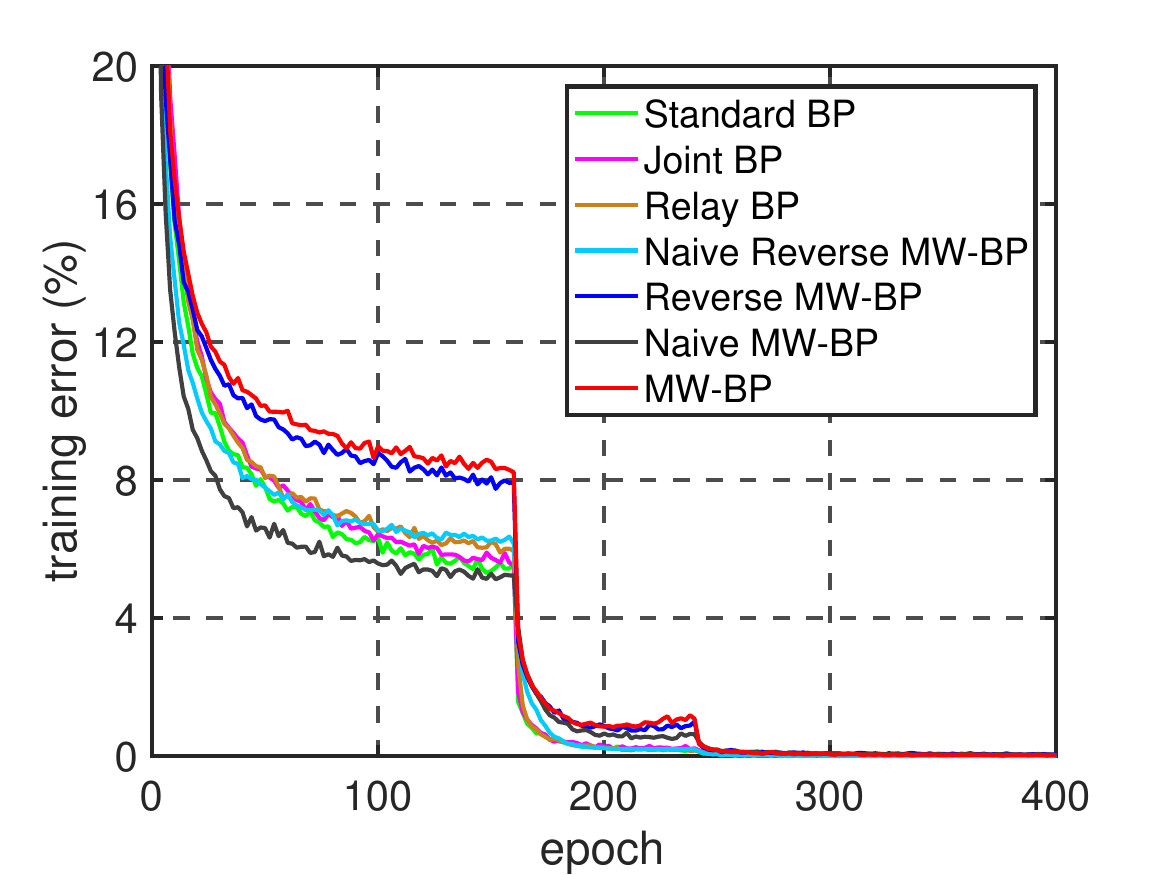}
		\label{fig:bptrain}
	}
	\caption{Performance comparison of different BP methods on CIFAR-10. (a) Testing error of different models. (b) Testing error of different BP methods on MwResNet-56-5. (c) Training error of of different BP methods on MwResNet-56-5.  All the curves come from the final outputs of deep models.
	}
	\label{fig:bpcompare}
\end{figure*}

The first variant is referred to as \textbf{Na\"{i}ve MW-BP}, in which a forward propagation is performed after each backpropagation. This method, however, may suffer from gradient vanishing issue since the deeper/final loss may decrease very quickly by updating the features and losses after each backpropagation. Therefore, the performance may be severely degraded (See Fig.~\ref{fig:bpcompare} and discussions in Section~\ref{sec:bpcomparison}). Moreover, multiple forward propagations will incur considerable training cost.

The second variant is referred to as \textbf{Reverse MW-BP}, in which we employ the same training paradigm \guo{of} MW-BP but conduct multiple backpropagations in the reverse order of MW-BP (\ie, from the last loss $\xi_K$ to the first loss $\xi_0$).
However, since the shallower models often have less representation power, the model update w.r.t. the shallower losses after the deeper losses may introduce representation errors into the whole network. In contrast, in MW-BP, the errors incurred by shallower losses can be corrected by deeper losses.

The third variant is referred to as \textbf{Na\"{i}ve Reverse MW-BP}. Based on {Reverse MW-BP},  we conduct a forward propagation after each backpropagation. 
However, in {Na\"{i}ve Reverse MW-BP}, since the shallower losses will not affect the deeper/final losses, the gradient vanishing may not be as severe as {Na\"{i}ve MW-BP}. 
Nevertheless, its performance is still limited since the model update w.r.t. shallower losses after deeper losses may bring in representation errors and hamper the overall performance.
Moreover, the multiple forward propagations will incur considerable training cost.

\subsection{Training and Inference Complexity}\label{sec:complexity}

Here, we investigate both the training and inference complexity \guo{of the proposed} MW-BP method.

\subsubsection{Training complexity}

The training cost of MW-BP is \aje{approximately} ($K/2+1$) times of the standard BP method since it conducts one \guo{shared} forward propagation and ($K+1$) backpropagations at each iteration.
If adding too many outputs to the model, it will greatly slow down the training process. In practice, introducing up to 4 auxiliary outputs is sufficient and is able to effectively improve the performance.
\guo{Although MW-BP has larger training cost than the standard BP method, it simultaneously produces ($K+1$) models.
Therefore, the increased complexity is acceptable when considering the cost for model selection.
% and provides an opportunity for model selection.
% In this sense, Multi-way BP can be more efficient.
}
% \subsection{Training Mw+Model with Existing Backpropagation Methods}

\subsubsection{Inference complexity} During the inference, we do not need to consider the auxiliary losses.  Moreover, MW-BP often produces very compact models (See results and discussions in Section~\ref{sec:compression}).  Thus, given the same architecture,  the models trained by MW-BP have the same inference cost to the ones trained by existing BP methods, but often exhibit better prediction performance.
In other words, under similar prediction performance, the inference cost for the models trained by MW-BP can be much lower than the ones trained with the standard BP method.

\section{Experiments}\label{sec:exp}
To demonstrate the effectiveness of MW-BP,
we apply the proposed MW-BP method to {various} architectures, including ResNet~\cite{he2015deep}, ResNeXt~\cite{xie2017aggregated}, DenseNet~\cite{huang2016densely}, Inception networks~\cite{DBLP:journals/corr/SzegedyIV16}, and MobileNet~\cite{sandler2018mobilenetv2}.
% \red{MobileNet~\cite{MobileNet}.}
For convenience, we use ``Mw+Architecture" to represent the model trained by MW-BP and ``Architecture-$L$-$\widehat{K}$'' to represent the model with $L$ layers and $\widehat{K}$ outputs, \eg, MwResNet-56-5. 
We {conduct experiments} on two tasks, namely \textbf{image classification} and \textbf{face recognition}.
All implementations are based on PyTorch.\footnote{The source code of MW-BP and the pretrained models are available at \href{https://github.com/tanmingkui/multiwaybp}{https://github.com/tanmingkui/multiwaybp}.}

We organize the experiments as follows.
{First}, {we study and compare different BP methods in Section~\ref{sec:bpcomparison}.}  {Second},  we extensively evaluate the proposed MW-BP method on image classification tasks in Section~\ref{sec:image_classification}.
{Third}, we compare the resultant models obtained by MW-BP with the \guo{light} models obtained by several compression methods in Section~\ref{sec:compression}.
{Fourth}, we 
% apply the MW-BP method to compact architectures like MobileFaceNet~\cite{chen2018mobilefacenets} and investigate the performance 
evaluate our method
on face recognition tasks in Section \ref{sec:face}.
% {{Moreover}, we conduct further experiments for the proposed method as follows.}

\subsection{Comparison of Various Backpropagation Methods}\label{sec:bpcomparison}
We compare MW-BP with 6 baseline BP methods, including the standard BP, Joint BP, Relay BP,
Na\"{i}ve MW-BP, Reverse MW-BP, and Na\"{i}ve Reverse MW-BP.
% Based on different architectures, we use the same number of auxiliary losses for these training methods.
% For Joint BP, we keep the same setting as that in~\cite{Lee2015}.
\guo{In this experiment, we first demonstrate the superiority of MW-BP over the standard BP. Then, we compare MW-BP with two existing BP methods that exploit auxiliary losses, namely Joint BP and Relay BP. Last, we compare MW-BP with its three variants. 
% We conduct these experiments on CIFAR-10~\cite{krizhevsky2009learning}.
}

\vspace{3 pt}
\noindent \textbf{Data Sets and Implementation Details.}
We compare the performance of different BP methods on CIFAR-10~\cite{krizhevsky2009learning}.
For Relay BP, we use the same setting in~\cite{shen2015learning} and set $c=1$, \ie, any intermediate layer only receives the gradients from the nearest two losses.
For all the considered BP methods, we use SGD to train the models for 400 epochs with a mini-batch size of 128. We initially set the learning rate to 0.1 and divide it by 10 at 40\% and 60\% of the total epochs.
{In this experiment, we set the weighting scalar to $\nu=2$ (See more discussions in Section~\ref{exp:nu}).}

\subsubsection{Comparison with the standard BP}
We apply MW-BP to the ResNet models with different depths and different numbers of auxiliary losses. We monitor the evolution of testing error based on different models in Fig.~\ref{fig:cifar10test}. 

\guo{From Fig.~\ref{fig:cifar10test}, MwResNet models consistently outperform their ResNet counterparts (of the same depth) trained by the standard BP, \eg, MwResNet-44-2 and MwResNet-56-2.
Moreover, when we increase the number of auxiliary losses, MwResNet-44-3 and MwResNet-56-5 achieve larger improvement than MwResNet-44-2 and MwResNet-56-2. 
Since the intermediate layers are able to receive sufficient supervision from the nearby losses (See Section~\ref{sec:multiwaybp}), MW-BP can effectively address the supervision vanishing issue and greatly improve the performance.}
However, the standard BP often incurs severe supervision vanishing issue and yields poor performance. 
% Specifically, as shown in Eqn.~(\ref{eq:backpropagation}), the transformation $\prod\nolimits_{j = l}^{L - 1} \bT_j(\bW_j)$ brought by the long backpropagation path makes the shallow layers very difficult to train. 
% As a result, the standard BP often incurs severe model redundancy.
These results strongly demonstrate the superiority of MW-BP over the standard BP method.

\begin{table*}[t!]
	\centering
	\caption{Performance comparison on CIFAR-10 and CIFAR-100. Note that \{$\cdot$/$N$\} indicates an N-fold increase in network width, \eg, WideResNet-28-2/10 is 10 times wider than the baseline model. ``-'' denotes the results that are not reported.}
	  \resizebox{0.8\textwidth}{!}
	  {
	\begin{tabular}{c|c|c|c|c|c|c}
		\hline
		\multirow{2}[0]{*}{Model} & \multicolumn{6}{c}{Error (\%)} \\
		\cline{2-7}
		&   \multicolumn{3}{c|}{CIFAR-10} & \multicolumn{3}{c}{CIFAR-100} \\
		\hline
        ResNet-20~\cite{he2016identity} & \multicolumn{3}{c|}{7.76} & \multicolumn{3}{c}{31.12} \\
        ResNet-32~\cite{he2016identity} & \multicolumn{3}{c|}{6.81} & \multicolumn{3}{c}{29.74} \\
        ResNet-44~\cite{he2016identity} & \multicolumn{3}{c|}{6.37} & \multicolumn{3}{c}{28.85} \\
        ResNet-56~\cite{he2016identity} & \multicolumn{3}{c|}{6.08} & \multicolumn{3}{c}{28.46} \\
        ResNet-110~\cite{he2016identity} & \multicolumn{3}{c|}{5.86} & \multicolumn{3}{c}{27.41} \\
        \hline
% 		FitNet~\cite{romero2014fitnets} & \multicolumn{3}{c|}{8.39} & \multicolumn{3}{c}{35.04} \\
        VGG-16~\cite{simonyan2014very} & \multicolumn{3}{c|}{6.01} & \multicolumn{3}{c}{27.07} \\
		DSN~\cite{Lee2015} & \multicolumn{3}{c|}{7.97} & \multicolumn{3}{c}{34.57} \\
		GoogLeNet~\cite{szegedy2015going} & \multicolumn{3}{c|}{-} & \multicolumn{3}{c}{21.97} \\
% 			Frac.Pool, 1 test~\cite{graham2014fractional} & \multicolumn{3}{c|}{4.50} & \multicolumn{3}{c}{31.20} \\
		Highway Network~\cite{srivastava2015highway} & \multicolumn{3}{c|}{7.60} & \multicolumn{3}{c}{32.24} \\
		StochResNet-110~\cite{DBLP:journals/corr/HuangSLSW16} & \multicolumn{3}{c|}{5.23} & \multicolumn{3}{c}{24.58} \\
% 		ResNet of ResNet/4~\cite{zhang2016residual} & \multicolumn{3}{c|}{3.77} & \multicolumn{3}{c}{19.73} \\
% 			SVB-Wide ResNet/10~\cite{jia2016improving} & \multicolumn{3}{c|}{3.58} & \multicolumn{3}{c}{18.32} \\
% 			ResNet-110~\cite{he2016identity} & \multicolumn{3}{c|}{5.86} & \multicolumn{3}{c}{27.41} \\
% 		ResNet-1001~\cite{he2016identity} & \multicolumn{3}{c|}{4.62} & \multicolumn{3}{c}{22.71} \\
% 			DARTS~\cite{liu2018darts} & \multicolumn{3}{c|}{5.93} & \multicolumn{3}{c}{25.64} \\
        MobileNetV2~\cite{sandler2018mobilenetv2} & \multicolumn{3}{c|}{8.35} & \multicolumn{3}{c}{28.33} \\
		WideResNet-28/10 ~\cite{zagoruyko2016wide} & \multicolumn{3}{c|}{4.17} & \multicolumn{3}{c}{20.50} \\
        ResNeXt-29~\cite{xie2017aggregated} & \multicolumn{3}{c|}{4.25} & \multicolumn{3}{c}{21.02} \\
        DenseNet-100~\cite{huang2016densely} & \multicolumn{3}{c|}{3.74} & \multicolumn{3}{c}{19.25} \\
		\hline
		\multicolumn{1}{c|}{Model-Depth-$\widehat K$}   & Joint BP & Relay BP & MW-BP & Joint BP & Relay BP & MW-BP \\
		\hline
        ResNet-20-2 & \multicolumn{2}{c|}{\textbf{7.53}} & 7.57 & \multicolumn{2}{c|}{30.67} & \textbf{30.54} \\
        ResNet-32-2 & \multicolumn{2}{c|}{6.73} & \textbf{6.54} & \multicolumn{2}{c|}{28.44} & \textbf{28.24} \\
        ResNet-44-2 & \multicolumn{2}{c|}{6.21} & \textbf{6.05} & \multicolumn{2}{c|}{28.03} & \textbf{27.46} \\
        % ResNet-44-3 & \multicolumn{2}{c|}{6.07} & \textbf{5.85} & \multicolumn{2}{c|}{27.86} & \textbf{27.19} \\
        ResNet-56-2 & \multicolumn{2}{c|}{5.96} & \textbf{5.77} & \multicolumn{2}{c|}{27.73} & \textbf{26.83} \\
        \cline{2-7}
        ResNet-44-3 & 6.07 & 6.03 & \textbf{5.85} & 27.86 & 27.77 & \textbf{27.19} \\
        ResNet-56-3 & 5.93 & 5.89 & \textbf{5.68} & 27.69 & 27.54 & \textbf{26.77} \\
        ResNet-56-5 & 5.83 & 5.77 & \textbf{5.53} & 27.37 & 27.33& \textbf{26.62} \\
        ResNet-110-5 & 5.62 & 5.57 &  \textbf{5.41}    & 26.94 & 26.88 & \textbf{26.48} \\
        % \hline
% 			DARTS-3 & 5.33 & 5.21 &  \textbf{5.04}    & 26.37 & 26.31 & \textbf{22.09} \\
        MobileNetV2-3 & 8.06  & 7.91 & \textbf{7.63} & 27.36 & 27.19 & \textbf{26.77} \\
		WideResNet-28-2/10 & 3.91 & 3.97 & \textbf{3.77} & 20.17 & 20.05 & \textbf{19.69} \\
		ResNeXt-29-3 & 3.98 & 3.84 &  \textbf{3.71}    & 19.23 & 19.17 & \textbf{18.96} \\
		DenseNet-100-4 & 3.65 & 3.61 & \textbf{3.53} & 19.21 & 19.24 & \textbf{19.13} \\
% 			MwResNet-2/10    & 3.48 & 3.41 & \textbf{3.32} & 18.53 & 18.19 & \textbf{17.87} \\
		\hline
	\end{tabular}
	  }
	\label{tab:cifar}%
\end{table*}

\subsubsection{Comparison of the BP methods with auxiliary losses}
We compare MW-BP with the two existing BP methods that exploit auxiliary losses to train deep models.
Based on ResNet-56, we {evenly} introduce 4 auxiliary losses at intermediate layers (at layer 15, 25, 35, 45, respectively). 
We compare the evolution of testing error and training error for different BP methods in Figs.~\ref{fig:bptest} and \ref{fig:bptrain}, respectively.

From Fig.~\ref{fig:bptest}, 
\guo{the proposed MW-BP significantly outperforms Joint BP and Relay BP and yields the best testing error of $5.53\%$. The main reason is that, unlike Joint BP and Relay BP, MW-BP conducts a backpropagation for each loss and does not sum up all the losses. In this way, MW-BP effectively avoids the loss conflict issue (See Section~\ref{sec:characteristics} for details).} 
However, Joint BP and Relay BP sum up all the losses and would inevitably incur the loss conflict issue. As a result, the proposed MW-BP method is able to obtain significantly better results than these methods.

\begin{table*}[t!]
	\centering
	\caption{Comparison with different models on ImageNet-2012 in terms of validation error  ({10-crop}).}
	\resizebox{0.79\textwidth}{!}
    {
		\begin{tabular}{c|c|c|c|c|c|c}
			\hline
			Model & \multicolumn{3}{c|}{Top-1 Error (\%)} & \multicolumn{3}{c}{Top-5 Error (\%)} \\
			\hline
			VGG-16~\cite{simonyan2014very} & \multicolumn{3}{c|}{28.07} & \multicolumn{3}{c}{9.33} \\
			GoogLeNet~\cite{szegedy2015going} & \multicolumn{3}{c|}{-} & \multicolumn{3}{c}{9.15} \\
% 			PReLU-net~\cite{he2015delving} & \multicolumn{3}{c|}{24.27} & \multicolumn{3}{c}{7.38} \\
% 			\hline
			ResNet-18~\cite{he2015deep} & \multicolumn{3}{c|}{28.43} & \multicolumn{3}{c}{9.97} \\
			ResNet-34~\cite{he2015deep} & \multicolumn{3}{c|}{24.76} & \multicolumn{3}{c}{7.35} \\
			ResNet-50~\cite{he2015deep} & \multicolumn{3}{c|}{22.85} & \multicolumn{3}{c}{6.71} \\
			ResNet-101~\cite{he2015deep} & \multicolumn{3}{c|}{21.75} & \multicolumn{3}{c}{6.05} \\
			Inception-ResNet~\cite{DBLP:journals/corr/SzegedyIV16} & \multicolumn{3}{c|}{18.77} & \multicolumn{3}{c}{4.13} \\
			\hline
			\multicolumn{1}{c|}{Model-Depth-$\widehat K$}  & Joint BP & Relay BP & MW-BP & Joint BP & Relay BP & MW-BP\\
			\hline
			ResNet-18-2 & \multicolumn{2}{c|}{28.13} & \textbf{27.70} & \multicolumn{2}{c|}{9.93} & \textbf{9.54} \\
			ResNet-34-2 & \multicolumn{2}{c|}{24.19} & \textbf{23.76} & \multicolumn{2}{c|}{7.26} & \textbf{7.03} \\
			ResNet-50-2 & \multicolumn{2}{c|}{22.78} & \textbf{22.47} & \multicolumn{2}{c|}{6.65} & \textbf{6.27} \\
			\cline{2-7}
			ResNet-50-4 & 22.64 & 22.57 & \textbf{22.15} & 6.46 & 6.24 & \textbf{6.07} \\
			ResNet-101-4 & 21.54 & 21.43 & \textbf{20.95} & 5.71 & 5.97 & \textbf{5.25} \\
			Inception-ResNet-4 & 18.75 & 18.71 & \textbf{18.61} & 4.15 & 4.10 & \textbf{4.05} \\
			\hline
		\end{tabular}
	}
	\label{tab:imagenet1000}%
\end{table*}

\subsubsection{Comparison with MW-BP variants}
We also compare MW-BP with its three variants to show the importance of the updating order of backpropagations and the shared forward propagation. 
% From Fig.~\ref{fig:bptest}, MW-BP consistently outperforms all the variants in terms of testing performance. 
% Detailed comparisons are as follows.

\guo{From Fig.~\ref{fig:bptest}, Na\"{i}ve MW-BP yields the worst testing performance among all the considered methods. However, the training error decreases very quickly at the beginning epochs (See Fig.~\ref{fig:bptrain}). With the decreased training error/loss, the gradient vanishing issue can be very severe and hamper the performance.
% Moreover, the errors incurred by the shallow losses would be propagated to the deep layers (See analysis in Section~\ref{sec:characteristic_share}). 
As a result, Na\"{i}ve MW-BP yields severely degraded performance.}

For Reverse MW-BP, when we reverse the updating order of MW-BP, it yields worse results than MW-BP. 
\guo{The main reason is that the model update w.r.t. the shallower losses after the deeper losses would introduce errors into the model (See discussions in Sections~\ref{sec:characteristic_order} and \ref{sec:variants}). As a result, the reverse updating order would hamper the overall performance (See Fig.~\ref{fig:bptest}).}
% However, since the separate backpropagation paradigm effectively avoid the loss conflict issue, Reverse MW-BP still significantly outperforms existing BP methods.

For Na\"{i}ve Reverse MW-BP, \guo{it adopts the reverse updating order of backpropagations and updates the features and losses before each backpropagation.}
% reverses the updating order of {Na\"{i}ve MW-BP} and conduct backpropagations from the last loss $\xi_K$ to the shallowest loss $\xi_0$. 
In this way, the shallow losses will not affect the deep losses and it significantly outperforms {Na\"{i}ve MW-BP} (See Fig.~\ref{fig:bptest}). However, 
% the update w.r.t. shallower losses after deeper losses may still introduce representation errors into the network. 
unlike MW-BP, Na\"{i}ve Reverse MW-BP with the reverse order cannot correct the errors incurred by shallow losses (See discussions in Section~\ref{sec:variants}). As a result, Na\"{i}ve Reverse MW-BP still yields slightly worse results than the proposed MW-BP method in Fig.~\ref{fig:bpcompare}.
% Moreover, the multiple forward propagations greatly increase the training cost. 
% From Fig.~\ref{fig:bptest}, it yields slightly worse resutls than MW-BP but greatly increases the training cost due to the multiple forward propagations.
% better results (See Fig.~\ref{fig:bptest}) and greatly reduces the risk of gradient vanishing issue, \guo{\ie, with relatively higher training error in Fig.~\ref{fig:bptrain}}.
% can be greatly reduced and all the layers are able to receive more sufficient supervision, resulting in significantly better performance than {Na\"{i}ve MW-BP}. 
% However, this method performs multiple forward propagations and greatly increases the training cost. 

% \guo{The above comparisons between MW-BP and its three variants demonstrate the importance of the updating order and shared forward propagation in MW-BP.}

% Unlike these methods, with the sequential updating order, MW-BP effectively avoids the representation error caused by shallow losses. In addition, the shared forward propagation also preserves the discriminative power of all the losses to improve the training. As a result, MW-BP is able to significantly outperform the considered BP methods (See Fig.~\ref{fig:bpcompare}).
% These results demonstrate the importance of the updating order and shared forward propagation in MW-BP.

\begin{table*}[t]
	\centering
	\caption{Comparison between the resultant MwResNet models and the \guo{light} models obtained by several state-of-the-art pruning methods on CIFAR-10.
% 	In column-2, ``50\%'' denotes the model with 50\% channels pruned.
    ResNet-110 is adopted as the baseline model.}
	\resizebox{1\textwidth}{!}{
		\begin{tabular}{c|c|c|cccccc|c}
			\hline
			\multicolumn{2}{c|}{Model} & Baseline &  PFEC~\cite{li2016pruning} & ThiNet~\cite{luo2017thinet} &
			CP~\cite{he2017channel} &
% 			FPGM~\cite{he2019filter} &
% 			MIL~\cite{dong2017more} &
			SFP~\cite{he2018soft} &
			PSFP~\cite{he2018progressive} &
			NISP~\cite{yu2018nisp} &
% 			DCP~\cite{zhuang2018discrimination} &
			MwResNet-56-5 \\
			\hline
			\multirow{2}[0]{*}{ResNet-110} & \#Params (M) & 1.73  & 1.16 & 0.87 & 0.87 &  1.10 & 1.10 & 0.98  & \textbf{0.85} \\
			\multirow{2}[0]{*}{on CIFAR-10} & \#FLOPs (M) & 253   & 155 & 127 & 127 &  150 & 150 & 143  & \textbf{127} \\
			& Error (\%)  & 5.86 &  6.70 & 6.22 & 5.91 & 5.83 & 6.06 & 6.04  & \textbf{5.53} \\
			\hline
		\end{tabular}
	}
	\label{tab:compression_cifar}
\end{table*}

\begin{table*}[t]
	\centering
	\caption{Comparison between the resultant MwResNet models and the \guo{light} models obtained by several state-of-the-art pruning methods on ImageNet (10-crop).
% 	In column-2, ``50\%'' denotes the model with 50\% channels pruned.
    ResNet-101 is adopted as the baseline model.}
	\resizebox{1\textwidth}{!}
	{
		\begin{tabular}{c|c|c|cccc|c}
			\hline
			\multicolumn{2}{c|}{Model} & Baseline & Rethinking~\cite{ye2018rethinking} & Taylor-FO-BN~\cite{molchanov2019importance} & SFP~\cite{he2018soft} & PSFP~\cite{he2018progressive}  & MwResNet-50-4 \\
			\hline
			\multirow{2}[0]{*}{ResNet-101} & \#Params (M) & 44.55 & \textbf{22.67} & 26.75 & 25.23 & 25.23  & 25.55  \\
			\multirow{2}[0]{*}{on ImageNet} & \#FLOPs (M) & 7260  & 3847 & 4790 & 4159  & 4159  & \textbf{3530}  \\
			& Top-1 Error (\%)  & 21.75 & 23.85 & 23.74 & 22.49 &  22.72  & \textbf{22.15} \\
			\hline
		\end{tabular}
	}
	\label{tab:compression_imagenet}
\end{table*}

% \begin{figure}[t]
% 	\centering
% 	\includegraphics[width = 0.9\columnwidth]{best_result_cifar10_test.eps}
% 	\caption{Performance comparison of different ResNet and MwResNet models in terms of testing error on CIFAR-10. All the curves come from the final outputs of deep networks.}
% 	\label{fig:cifar10test}
% \end{figure}

\subsection{Experiments on Image Classification}\label{sec:image_classification}

We apply the proposed MW-BP method to various architectures,
including ResNet~\cite{he2015deep}, WideResNet~\cite{zagoruyko2016wide}, DenseNet~\cite{huang2016densely}, MobileNet~\cite{sandler2018mobilenetv2} and Inception-ResNet~\cite{DBLP:journals/corr/SzegedyIV16}. In this experiment, we evaluate our method on several image classification data sets.
% We compare the performance of different methods on several benchmark data sets, including CIFAR-10~\cite{krizhevsky2009learning}, CIFAR-100~\cite{krizhevsky2009learning} and ImageNet~\cite{russakovsky2015imagenet}.

\subsubsection{Compared Methods}

% \guo{For convenience, we use model-$L$-$\widehat{K}$ to denote the model with $L$ layers and $\widehat{K}$ outputs, \eg, ResNet-56-5.}
{We compare the proposed MW-BP with three existing BP methods, including the standard BP, Joint BP, and Relay BP.}
\guo{Moreover, we also consider several state-of-the-art deep learning models for comparison. On CIFAR-10 and CIFAR-100, we compare the models trained by MW-BP with VGG~\cite{simonyan2014very}, GoogLeNet~\cite{szegedy2015going}, DSN~\cite{Lee2015}, StochResNet~\cite{DBLP:journals/corr/HuangSLSW16}, etc.
% FitNet~\cite{romero2014fitnets}, , 
On ImageNet, we compare our models with several widely used deep models, namely VGG, GoogLeNet, ResNet, and Inception-ResNet.}
% Several state-of-the-art deep learning models are considered for comparison, including VGG~\cite{simonyan2014very}, GoogLeNet~\cite{szegedy2015going}, FitNet~\cite{romero2014fitnets}, Highway Network~\cite{srivastava2015highway}, StochResNet~\cite{DBLP:journals/corr/HuangSLSW16}, etc.
% \guo{To demonstrate the effectiveness of Multi-way BP in obtaining compact models, we compare the resultant models trained by Multi-way BP with the compressed models by several channel pruning methods, including CP~\cite{he2017channel}, ThiNet~\cite{luo2017thinet}, NISP~\cite{yu2018nisp}, PFEC~\cite{li2016pruning}, FPGM~\cite{he2019filter}, DCP~\cite{zhuang2018discrimination}, etc.}
%
% Rethinking~\cite{ye2018rethinking}, Taylor-FO-BN~\cite{molchanov2019importance}, SFP~\cite{he2018soft}, and PSFP~\cite{he2018progressive}.

\subsubsection{Data Sets and Implementation Details}
We conduct comparisons on {several benchmark data sets}, including CIFAR-10~\cite{krizhevsky2009learning}, CIFAR-100~\cite{krizhevsky2009learning}, and ImageNet~\cite{russakovsky2015imagenet}.
% CIFAR-10 contains 10 classes of 32x32 natural color images, each with 5,000 training samples
% and 1,000 testing samples. CIFAR-100 contains 100 classes. Each class has 500 training samples and 100 testing samples.
% SVHN contains house number images of 32x32 pixels, which includes 73,237 digits for training, 26,032 digits for testing, and 531,131 digits as extra training data.
% ImageNet contains 1,000 classes with 1.28 million training images and 50k testing images.
On CIFAR-10 and CIFAR-100, we perform SGD with a mini-batch size of 128 and train the model for 400 epochs. The learning rate starts from 0.1 and is divided by 10 at 40\% and 60\% of total epochs.
On ImageNet, we use a mini-batch size of 256. For each model, we use the same number of epochs and the same learning rate strategy as the original paper. Specifically, we train ResNet models for 90 epochs and Inception models for 160 epochs.
% All model parameters are initialized as in~\cite{he2015delving}.
{In all experiments, we empirically set the weighting scalar to $\nu=2$ (See discussions and results in Section~\ref{exp:nu}).}

\subsubsection{Comparison on CIFAR-10 and CIFAR-100}\label{sec:cifar}
% We perform 400 SGD epochs for the training, and use the same data augmentation as in~\cite{Lee2015}.
% In this experiment, we first compare the testing error evolution of
% 2 ResNets with (44, 56)-layers and 4 MwResNets of different settings in Fig.~\ref{fig:cifar10test}.
% Then, we comprehensively compare MW-BP with other training methods in Table~\ref{tab:cifar}. 

% \noindent \emph{\textbf{Comparisons with state-of-the-art models}}.

We conduct a comprehensive comparison between {the proposed MW-BP method} and existing BP methods based on various architectures, including ResNet~\cite{he2015deep}, Wide ResNet~\cite{zagoruyko2016wide}, ResNeXt~\cite{xie2017aggregated}, DenseNet~\cite{huang2016densely}, and MobileNet~\cite{sandler2018mobilenetv2}. We show the comparison results on CIFAR-10 and CIFAR-100 data sets in Table~\ref{tab:cifar}.
% {In this experiment, we apply different BP methods to the following deep architectures, including
% ResNet~\cite{he2015deep}, Wide ResNet~\cite{zagoruyko2016wide}, ResNeXt~\cite{xie2017aggregated}, and DenseNet~\cite{huang2016densely}.}

\guo{From Table \ref{tab:cifar}, we have the following observations. {First}, the models trained by MW-BP significantly outperform the models trained by existing BP methods. For example, MwResNet-56-5 yields much better performance than the ResNet-56 baseline model trained by the standard BP and the ResNet-56-5 counterparts trained by Joint BP and Relay BP. Second, the proposed MW-BP is able to effectively reduce the model redundancy and produce compact models. To be specific, MwResNet-44 with 44 layers yields comparable or even better results than ResNet-110 with 110 layers on both CIFAR-10 and CIFAR-100 data sets. 
{Third}, when we introduce more auxiliary losses, we can further improve the performance. For example, MwResNet-56-5 with 5 losses yields better results than MwResNet-56-2 with 2 losses.}

\guo{Besides the ResNet models, we also apply the proposed MW-BP to several state-of-the-art architectures, such as ResNeXt, DenseNet, and MobileNet. From Table \ref{tab:cifar}, the resultant models trained by MW-BP consistently outperform the models trained by existing BP methods based on various architectures. For ResNeXt and DenseNet, MW-BP yields the best performance among all the considered BP methods. Even for a very compact model MobileNetV2, our MwMobileNetV2-3 also significantly outperforms the models trained by other BP methods.
These results demonstrate that the proposed MW-BP method exhibits good compatibility with the considered deep architectures.
}
% and shows good compatibility with these architectures.
% It is worth mentioning that the proposed Multi-way BP method can be easily applied to most deep architectures with slight modifications.

\begin{figure*}[t]
	\centering
	\subfigure[Validation error on LFW.]{
		\includegraphics[trim = 2mm 0mm 12mm 5mm,
		clip, width = 0.65\columnwidth]{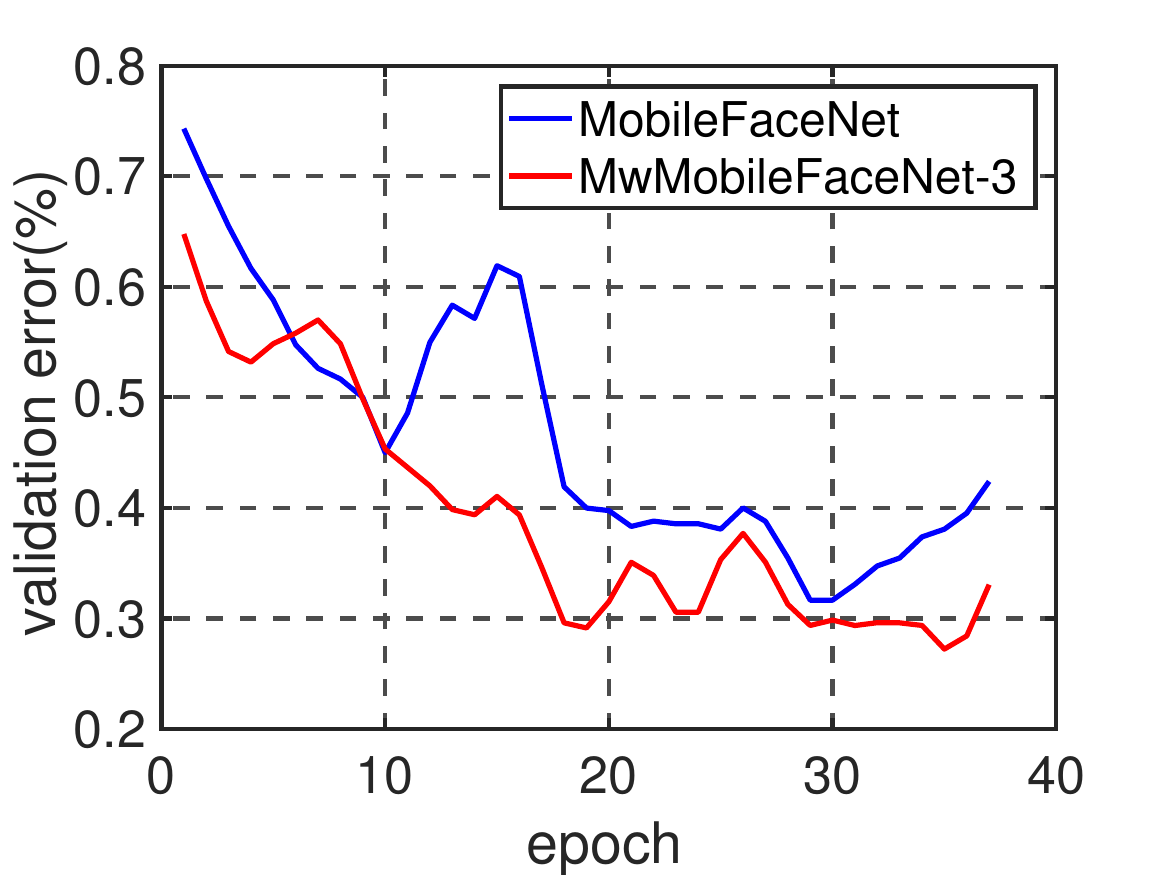}
	}
	\subfigure[Validation error on CFP-FP.]{
		\includegraphics[trim = 2mm 0mm 12mm 5mm,
		clip, width = 0.65\columnwidth]{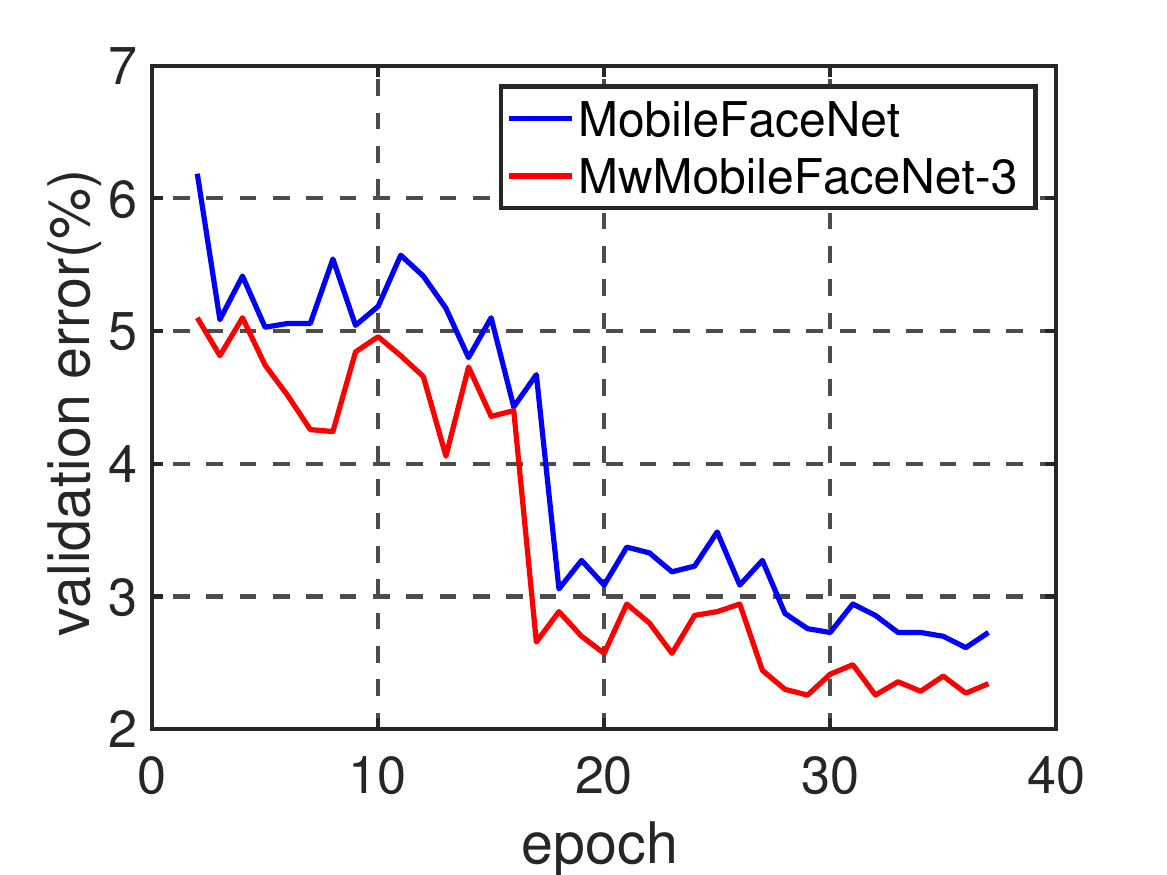}
	}
	\subfigure[Validation error on AgeDB-30.]{
		\includegraphics[trim = 2mm 0mm 12mm 5mm,
		clip, width = 0.65\columnwidth]{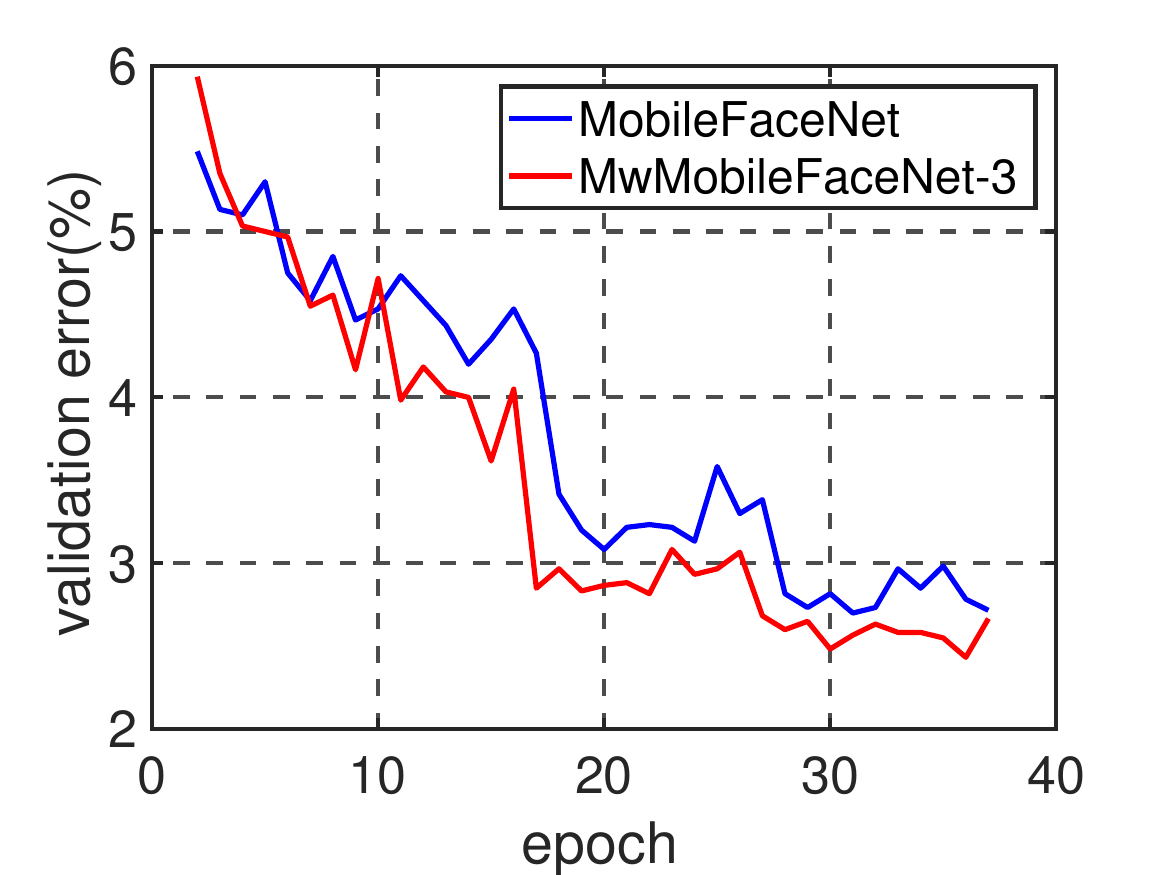}
	}
	\caption{Performance comparison of the MobileFaceNet models trained with and without MW-BP on 3 benchmark face recognition data sets.}
	\label{fig:faceplot}
\end{figure*}

\subsubsection{Comparison on ImageNet}\label{sec:imagenet}
We also evaluate the proposed method on a large-scale data set ImageNet~\cite{krizhevsky2012imagenet}.
\guo{In this experiment, we apply the MW-BP method to several widely used models, \eg, ResNet and Inception network.}
% We follow the same experimental settings in~\cite{he2015deep}.
% We scale the learning rate by a factor of 0.5 to train the models.
% {For fair comparison, we train deep models with the same number of epochs as that in the original paper. Specifically, we train ResNet models for 90 epochs and Inception models for 160 epochs.}
We show the comparison results in Table~\ref{tab:imagenet1000}.

From Table~\ref{tab:imagenet1000}, the proposed MW-BP method consistently outperforms existing BP methods based on various models.
% the standard BP, Joint BP, and Relay BP.
% achieves better performance than ResNet.
For example, MwResNet-18-2, MwResNet-34-2 and MwResNet-50-2 with two outputs yield better results than the models trained by the standard BP, Joint BP, and Relay BP.
% with the same number of layers.
When adding more auxiliary losses, MW-BP is able to obtain better performance, \eg, MwResNet-50-4.
% to 22.15\% and 6.07\% on top-1 error and top-5 error respectively.
% Moreover, Mw+Models trained with Multi-way BP consistently outperforms their Joint BP and Relay BP counterparts on a broad range of model designs.
We further apply MW-BP on large models like ResNet-101 and Inception-ResNet.
% and follow the same training setting as~\cite{DBLP:journals/corr/SzegedyIV16}.
% From Table~\ref{tab:imagenet1000},
When we increase the number of losses up to 4, our MwResNet-101-4 obtains a significant performance improvement of 0.8\% in terms of Top-1 error. Moreover, our MwInception-ResNet-4 trained by MW-BP yields the best performance among all the considered models in terms of both Top-1 error and Top-5 error. These results demonstrate the effectiveness of our MW-BP method.

\begin{table*}[htbp]
  \centering
  \caption{Performance comparison of different methods for face recognition. ``VR'' refers to Verification TAR (True Accepted Rate) and ``FAR$10^{-6}$'' refers to the False Accepted Rate at $10^{-6}$. ``-'' denotes the results that are not reported.}
  \resizebox{0.68\textwidth}{!}
{
    \begin{tabular}{c|c|c|c|c}
    \hline
    \multirow{2}[0]{*}{Model} & \multicolumn{3}{c|}{Validation Accuracy (\%)} & \multirow{1}[0]{*}{VR@FAR$10^{-6}$ (\%)} \\
    \cline{2-5}
     & \multicolumn{1}{c|}{LFW} & \multicolumn{1}{c|}{CFP-FP} & \multicolumn{1}{c|}{AgeDB-30} & MegaFace \\
    \hline
    SphereFace~\cite{liu2017sphereface} & 99.42 & - & - & 85.56 \\
    CosFace~\cite{wang2018cosface} & 99.33 & - & - & 89.88 \\
    % MobileFaceNet~\cite{chen2018mobilefacenets} & 99.55 & 91.80  & 95.43 & 90.41 \\
    MobileFaceNet~\cite{chen2018mobilefacenets} & {99.67} & {97.30}  & {97.02} & {93.57} \\
    MwMobileFaceNet-3 & \textbf{99.72} & \textbf{97.60}  & \textbf{97.45} & \textbf{93.94} \\
    \hline
    \end{tabular}%
    }
  \label{tab:face}%
\end{table*}%

\subsection{Comparison with Light Models}\label{sec:compression}
% With MW-BP, we can {effectively reduce the internal model redundancy and} obtain very compact models.
% \aje{that} outperform their very deep equivalents.
% In this sense, the shallow compact models often yield better or comparable performance but with much fewer parameters than their deep equivalents.
%In general, fewer model parameters imply less storage space and faster inference, which is important for real-world applications.
% In term of model compactness,
% To better illustrate this,
% we compare the resultant models with the models compressed by state-of-the-art channel pruning methods.
% including Weight Sum~\cite{li2016pruning}, ThiNet~\cite{luo2017thinet}, Channel Pruning (CP)~\cite{he2017channel}, NISP~\cite{yu2018nisp}, PFEC~\cite{li2016pruning}, FPGM~\cite{he2019filter}, and DCP~\cite{zhuang2018discrimination}.
% We adopt ResNet-110 as the baseline model and show the results in Table~\ref{tab:compression}.
To demonstrate the superiority of our method in model compactness, we compare the resultant models obtained by MW-BP with the deeper models
% (\eg, ResNet-110 on CIFAR-10 and ResNet-101 on ImageNet)
compressed by several channel pruning methods.
In this way, all the models have approximately the same model complexity.
We show the detailed comparison results in Tables~\ref{tab:compression_cifar} and~\ref{tab:compression_imagenet}.

{For the considered channel pruning methods, we adopt ResNet-110 and ResNet-101 as the baseline models on CIFAR-10 and ImageNet, respectively.}
From Tables~\ref{tab:compression_cifar} and~\ref{tab:compression_imagenet}, most pruning methods yield similar or worse performance than the baseline models.
% On CIFAR-10, two state-of-the-art pruning methods FPGM and DCP yield better performance than the baseline model and achieve approximately $2\times$ parameters reduction.
% For example, Weight Sum  keeps the same level of accuracy as the baseline but only obtains 1.03$\times$ model compression and 1.17$\times$ inference acceleration. For ThiNet~\cite{luo2017thinet} and Channel Pruning~\cite{he2017channel}, when we increase the acceleration rate to 2$\times$, both methods incur great accuracy drops and yield the errors of 6.34\% and 6.86\%, respectively.
% Compared to these methods,
% our MwResNet-44-3 achieves the same level of accuracy as ResNet-110 with 110 layers. More critically, MwResNet-44-3 has a relatively high compression rate with 2.62$\times$ FLOPs reduction and 2.59$\times$ parameters reduction.
Compared to the considered methods, the resultant model trained by MW-BP obtains the best or comparable results
% outperforms the models obtained by most channel pruning methods
in terms of both accuracy and model compactness.
For example, on CIFAR-10, MwResNet-56-5 with 56 layers and 5 outputs
% has the same compression rate (2$\times$ acceleration) as ThiNet, CP, and DCP, but
achieves a significant accuracy improvement of 0.33\% compared to the baseline ResNet-110 and yields a great reduction of model size.
On ImageNet, our MwResNet-50-4
% has approximately the same compression rate as the considered channel pruning methods and
yields the best performance with the smallest accuracy drop compared to the baseline ResNet-101.
These results show that
% by improving the representation ability of intermediate layers,
the proposed MW-BP method effectively is able to reduce the internal model redundancy and thus produces very compact models.
% 	than existing model compression methods.

\begin{table*}[t]
	\centering
	\caption{Testing error of the intermediate models obtained by different BP methods on CIFAR-10 and CIFAR-100.}
	\resizebox{1\textwidth}{!}
	{
		\begin{tabular}{c|c|c|c|c|c|c|c|c|c|c}
			\hline
			\multirow{2}[0]{*}{Model} & \multirow{2}[0]{*}{\#Layers} & \multirow{2}[0]{*}{\#Params} & \multicolumn{4}{c|}{CIFAR-10 Error (\%)} & \multicolumn{4}{c}{CIFAR-100 Error (\%)}  \\
			\cline{4-11}
			&       &       & Standard BP & Joint BP & Relay BP & MW-BP & Standard BP & Joint BP & Relay BP & MW-BP \\
			\hline
			model-15 & 15    & 0.03M & 63.01 & 61.37 & 58.93 & \textbf{50.35} & 87.74 & 84.51 & 79.85 & \textbf{71.73} \\
			model-25 & 25    & 0.09M & 45.07 & 40.11 & 39.47 & \textbf{18.94} & 68.17 & 63.88 & 60.37 & \textbf{51.21} \\
			model-35 & 35    & 0.18M & 34.01 & 28.92 & 27.64 & \textbf{9.23} & 49.54 & 43.17 & 41.09 & \textbf{35.88} \\
			model-45 & 45    & 0.48M & 13.71 & 11.56 & 10.21 & \textbf{5.67} & 35.72 & 31.63 & 30.44 & \textbf{27.35}  \\
			model-56 & 56    & 0.85M & 6.08 & 5.83 & 5.77 & \textbf{5.53} & 28.46 & 27.37 & 27.33 & \textbf{26.62} \\
			\hline
		\end{tabular}
	}
	\label{tab:intermediate}
\end{table*}

\subsection{Experiments on Face Recognition}\label{sec:face}

\guo{
% Our method has good generalization performance on other recognition tasks such as face recognition. 
In this experiment, we further apply MW-BP to a face recognition model MobileFaceNet~\cite{chen2018mobilefacenets}. We evaluate the models trained by MW-BP on several face recognition data sets.}

% {In this experiment, we apply our MW-BP method to very compact architectures, \eg, MobileNet~\cite{sandler2018mobilenetv2}.
% We evaluate our method on several face recognition tasks.
% }

\subsubsection{Compared Methods}
% Due to the remarkable performance of MobileNet~\cite{sandler2018mobilenetv2} on image classification, Chen \etal adapt it to face recognition and develop a new model named MobileFaceNet~\cite{chen2018mobilefacenets}. 
We apply the proposed MW-BP method to train MobileFaceNet in which we add 2 auxiliary losses at the intermediate layers. For convenience, we term it MwMobileFaceNet-3 (\ie, containing 3 losses in total).
We compare the resultant models with the baseline model trained by the standard BP.
% To show the superiority of the proposed MW-BP, 
% we compare the resultant model trained by the standard BP and 
Moreover, we also consider several face recognition models for comparison, including SphereFace~\cite{liu2017sphereface} and CosFace~\cite{wang2018cosface}.
% Besides MobileFaceNet, we also consider several face recognition models in the comparison, including SphereFace~\cite{liu2017sphereface} and CosFace~\cite{wang2018cosface}.

\subsubsection{Data Sets and Implementation Details}

\guo{We adopt the large-scale data set MS1M~\cite{guo2016ms} as the training data and four benchmark data sets as the validation data, including LFW~\cite{huang2008labeled}, CFP-FP~\cite{sengupta2016frontal}, AgeDB-30~\cite{moschoglou2017agedb}, and MegaFace~\cite{kemelmacher2016megaface}}.
% We train the model on a large-scale data set MS1M~\cite{guo2016ms}
% that contains 5.1M images of 93K identities in total.
% on a cleaned data set (containing 5.1M images of 93K identities)
% {In this experiment, we adopt MobileFaceNet as the baseline method. We add 2 auxiliary losses at the intermediate layers and term it MwMobileFaceNet-3 (\ie, containing 3 losses in total) for convenience.}
We follow the same setting as that in MobileFaceNet~\cite{chen2018mobilefacenets}. Specifically,
all face images are preprocessed to the size of $112 \times 112$.
We set the momentum and weight decay to 0.9 and $4 \times 10^{-5}$, respectively. We train the models for 36 epochs with a mini-batch size of 200. The learning rate starts from 0.1 and is divided by 10 at the \{15, 25, 31\}-th epoch, respectively.
% Different from image classification, we only need to train a general model for face recognition and evaluate it on different validation data sets.
{Following the setting of image classification, we use the same weighting scalar $\nu=2$ in the face recognition experiments.}

\subsubsection{Performance Comparison}

% {To show the effectiveness of MW-BP,}
In this experiment, we compare the models trained with and without MW-BP in terms of the evolution of validation error. Fig.~\ref{fig:faceplot} show the comparison on 3 data sets, including LFW, CFP-FP, and AgeDB-30. From Fig.~\ref{fig:faceplot}, the proposed MW-BP method greatly accelerates the convergence and yields significantly better performance than the standard BP method.

Besides the aforementioned 3 data sets, we also evaluate the models on a large-scale data set MegaFace and show more detailed results in Table~\ref{tab:face}. From Table~\ref{tab:face}, our MwMobileFaceNet-3 consistently outperforms the considered baseline models on 4 data sets. These results demonstrate the effectiveness of the proposed MW-BP method on face recognition models.

\begin{table*}[t]
  \centering
  \caption{Effect of weighting scalar $\nu$ on CIFAR-10. MwResNet-56-5 is adopted as the baseline model.}
%   \resizebox{0.7\textwidth}{!}
  {
    \begin{tabular}{c|c|c|c|c|c|c}
    \hline
    \multirow{2}[0]{*}{Model} & \multirow{2}[0]{*}{$\nu$}    & \multicolumn{5}{c}{Error (\%)} \\
    \cline{3-7}
    & & model-15 & model-25 & model-35 & model-45 & model-56 \\
    \hline
    \multirow{5}[0]{*}{MwResNet-56-5} & 0     &   57.63    &  25.30     &    15.94   &    11.89   & 8.43 \\
     \multirow{5}[0]{*}{(ResNet-56 6.08\%)}     & 1     &   49.83    &   18.77    &   9.71    &    5.97   & 5.90 \\
          & 2     & 50.35 & 18.94 & 9.23  & {5.67}  & \textbf{5.53} \\
          & 5     &   52.46    &   19.97    &  10.93     &  6.15     & 6.03 \\
        %   \hline
    \cline{2-7}
          & Adaptive-I & \textbf{41.47} & \textbf{16.05} & \textbf{8.65}  & 6.19  & 6.08 \\
          & Adaptive-II & {44.89} & {16.98} & {9.07}  & \textbf{5.64}  & {5.61} \\
    \hline
    \end{tabular}%
    }
  \label{tab:weight}%
\end{table*}%

\section{Further Experiments}\label{sec:discussion}

In this section, we investigate the prediction ability of intermediate models obtained by MW-BP and conduct ablation studies \guo{for the proposed method}.

%we conduct further experiments for the proposed MW-BP method. 

%In Section~\ref{sec:intermediate}, we first investigate the \guo{prediction ability of intermediate models} associated with the auxiliary losses.
% and the performance of safe prediction scheme in Section~\ref{sec:intermediate_safe}, respectively.
%Then,
% we perform ablation study to understand
%we \guo{study the effect of some hyperparameters} in MW-BP in Section \ref{sec:hyperparameters}.

\subsection{Prediction Ability of Intermediate Models}\label{sec:intermediate}
% Note that the MW-BP training process inherently produces multiple intermediate models of differing depths.
% The MW-BP inherently produces multiple intermediate models of differing depths.
In this section, we investigate the prediction ability of the intermediate models obtained during the training process of MW-BP.
In Table~\ref{tab:intermediate}, we compare the performance of intermediate models generated by each loss of MwResNet-56-5
against {the models trained by the standard BP}\footnote{To obtain the intermediate models of ResNet-56, We fix the parameters of all layers and only train the outputs added to the intermediate layers.}, Joint BP, and Relay BP.

From Table~\ref{tab:intermediate}, each intermediate model of MwResNet-56-5 consistently outperforms its competitors (of the same depth) obtained by the standard BP, Joint BP, and Relay BP.
% More critically, some intermediate models (\eg, ) even outperforms the full-depth ResNet counterpart, \guo{\eg, model-56}.
By comparing these results with the results in Table \ref{tab:cifar},
% we see that the MwResNet-56-5
the intermediate model
\textbf{model-45} {even outperforms very deep models.}
{For example, compared to ResNet-110 with 1.7M parameters (5.86\% error on CIFAR-10 and 27.41\% error on CIFAR-100), our \textbf{model-45} with 0.85M parameters yields better performance on both CIFAR-10 (5.67\% error) and CIFAR-100 (27.35\% error).}
% \guo{has fewer parameters of 0.48M} but
% yields much better performance ($5.67\%$ error on CIFAR-10 and 27.35\% error on CIFAR-100) even than ResNet-110 (5.86\% error on CIFAR-10 and 27.41\% error on CIFAR-100 {with 1.7M parameters}).
These results demonstrate that the proposed method not only improves the {representation ability} of intermediate layers, but also provides the opportunity for a form of model selection.

\subsection{Effect of the Weighting Scalar $\nu$}\label{exp:nu}
% We investigate how to weight outputs to achieve better performance.
We investigate the effect of the weighting scheme.
{Two kinds of weighting scheme are considered. First, we can set $\nu$ to a constant value to adjust the weights of different losses. Second, we also consider two adaptive strategies to dynamically increase or decrease the weights during the training, namely \emph{Adaptive-I} and \emph{Adaptive-II}. 
\guo{In \emph{Adaptive-I}, we initially set $\nu{=}1/2$ and multiply it by 2 when we change the learning rate.
% Finally, the weighting scalar ends up with $\nu=2$.
Just opposite to \emph{Adaptive-I}, in \emph{Adaptive-II}, we initially set $\nu{=}2$ and divide it by 2 along with the change of learning rate.}
Based on MwResNet-56-5, we compare the performance of the models trained with different weighting strategies in Table~\ref{tab:weight}.}

% Besides setting $\nu$ to a constant value, we also construct two adaptive strategies that adjust $\nu$ during the training process.
% Under the setting of $\nu$ described in Section~\ref{sec:weighting},
% we dynamically increase or decrease $\nu$ in the range of $[1/2, 2]$. For convenience, we term the two strategies Adaptive-I and Adaptive-II, respectively.
% In Adaptive-I, we initially set $\nu=1/2$ and multiply it by 2 when we change the learning rate (\ie, at 40\% and 60\% of total epochs, respectively). Finally, the weighting scalar ends up with $\nu=2$.
% Just opposite to Adaptive-I, in Adaptive-II, $\nu$ starts from 2 and is divided by 2 along with the change of learning rate, ending up with $\nu=1/2$.

{
We first compare the effect of $\nu$ with different constant values.
% In general, a smaller $\nu$ indicates the larger weights for the shallow losses.
When we set $\nu{=}0$, all the losses are equally weighted. Since multiple losses are not equally important (See Section~\ref{sec:weighting}), the equally weighted losses severely hamper the performance of MW-BP in Table~\ref{tab:weight}.}
% {Thus, we should set $\nu \geq 1$ to reduce the weights for shallow-layer losses.}
% due to their decreasing discriminative powers.
However, when we choose a large value of $\nu{=}5$, the weights would decay so aggressively that the effects of auxiliary outputs are negligible. To avoid this issue, we empirically choose $\nu{=}2$ and this setting yields the best performance in practice.

For the two adaptive strategies, from Table \ref{tab:weight}, they yield slightly worse results than the best setting of $\nu{=}2$ at the final output. However, they significantly improve the performance of intermediate models. Therefore, we suggest that one can use the adaptive strategies to obtain better intermediate models.
% These results demonstrate that the proposed weighting scheme effectively alleviates the conflicts among multiple losses.

% We suggest $\nu=1$ or $\nu=2$ to achieve the best performance.

\begin{table}[t]
  \centering
  \caption{Testing error of varying number of losses in MwResNet-56 on CIFAR-10.}
    % \resizebox{0.3\textwidth}{!}{
    \begin{tabular}{c|c|c}
    \hline
    Model & \#Losses & Error (\%) \\
    \hline
    ResNet-56~\cite{he2016identity}     & 1 & 6.08 \\
    \hline
    \multirow{4}{*}{MwResNet-56}
        & 2 &  5.77 \\
        & 5 &  \textbf{5.53} \\
        & 10 &  7.36 \\
        & 25 &  9.18 \\
    \hline
    \end{tabular}
    % }
  \label{tab:num_outputs}%
\end{table}%

\subsection{Effect of the Numbers of Losses} \label{exp:num_outputs}
We investigate the effect of the number of losses.
% Note that a large number of outputs often implys a small output interval, e.g. MwResNet-56-5 with $\tau=10$ while MwResNet-56-25 with $\tau=2$.
{We take ResNet-56 for example and insert different numbers of losses.}
% We show the results in Table~\ref{tab:num_outputs}.
From the results in Table~\ref{tab:num_outputs},
MwResNet-56-2 and MwResNet-56-5 perform significantly better than ResNet-56.
However, adding too many losses does not necessarily improve the performance. For example, the MwResNet-56 models with 10 and 25 outputs yield severely degraded performance.
% perform much worse than MwResNet-56-2, MwResNet-56-5 and ResNet-56.
% Note that increasing the number of outputs will decrease the output interval.
% In this case, the interval becomes very small $\tau=2$. As a result,
The main reason is that the interval between losses is too small so that they may affect each other and eventually degrade the performance.
% due to the overfitting issue.
Moreover, adding too many losses will also slow down the training.
% We thus suggest avoiding adding too many outputs in the networks.
In practice, introducing up to 5 losses is sufficient to effectively improve the performance according to previous experimental results.

\section{Conclusion}
In this paper, we have investigated the supervision vanishing issue in {existing backpropagation (BP) methods for} training deep networks. When the network \aje{is} very deep, shallow layers tend to receive insufficient supervision due to the {severe transformation} through long backpropagation path, resulting in great model redundancy. To address \aje{these issues},
{we introduced auxiliary losses into deep models and proposed an effective training method, called Multi-way BP (MW-BP).}
% \guo{that relies on multiple losses added to the intermediate layers of deep networks.}
% we developed a model called Mw+Models by introducing auxiliary losses at intermediate layers.
% To train the models more effectively, we propose the Multi-way backpropagation method to train the network with multiple losses.
Based on various architectures, our method consistently obtains significant performance improvement and produces more compact models than existing BP methods.
{More critically, with approximately the same model complexity, the resultant models also outperform the light models obtained by state-of-the-art model compression methods.}
Moreover, the proposed MW-BP method is able to inherently produce multiple intermediate models at the same time, which offers an opportunity for a form of model selection.
% By taking all the intermediate models, we further propose a new inference method to reduce the overfitting risk for a safer prediction.
% We apply the proposed MW-BP method to various architectures and conduct experiments on both image classification and face recognition tasks to show the effectiveness of our method. 
Extensive experiments {on both image classification and face recognition} tasks demonstrate the effectiveness of the proposed method.

\ifCLASSOPTIONcaptionsoff
  \newpage
\fi

{
\bibliographystyle{plain}
\bibliography{auxnet-full}

\begin{thebibliography}{10}

\bibitem{ba2014deep}
Jimmy Ba and Rich Caruana.
\newblock {D}o {D}eep {N}ets {R}eally {N}eed to be {D}eep?
\newblock In {\em Advances in Neural Information Processing Systems}, pages
  2654--2662, 2014.

\bibitem{DBLP:journals/pami/BadrinarayananK17}
Vijay Badrinarayanan, Alex Kendall, and Roberto Cipolla.
\newblock Segnet: {A} deep convolutional encoder-decoder architecture for image
  segmentation.
\newblock {\em IEEE Transactions on Pattern Analysis and Machine Intelligence},
  39(12):2481--2495, 2017.

\bibitem{DBLP:journals/pami/ChenPKMY18}
Liang{-}Chieh Chen, George Papandreou, Iasonas Kokkinos, Kevin Murphy, and
  Alan~L. Yuille.
\newblock Deeplab: Semantic image segmentation with deep convolutional nets,
  atrous convolution, and fully connected crfs.
\newblock {\em IEEE Transactions on Pattern Analysis and Machine Intelligence},
  40(4):834--848, 2018.

\bibitem{chen2018mobilefacenets}
Sheng Chen, Yang Liu, Xiang Gao, and Zhen Han.
\newblock Mobilefacenets: Efficient cnns for accurate real-time face
  verification on mobile devices.
\newblock In {\em Chinese Conference on Biometric Recognition}, pages 428--438,
  2018.

\bibitem{collobert2008unified}
Ronan Collobert and Jason Weston.
\newblock {A} {U}nified {A}rchitecture for {N}atural {L}anguage {P}rocessing:
  {D}eep {N}eural {N}etworks with {M}ultitask {L}earning.
\newblock In {\em International Conference on Machine Learning}, pages
  160--167, 2008.

\bibitem{DBLP:journals/pami/DollarABP14}
Piotr Doll{\'{a}}r, Ron Appel, Serge~J. Belongie, and Pietro Perona.
\newblock Fast feature pyramids for object detection.
\newblock {\em IEEE Transactions on Pattern Analysis and Machine Intelligence},
  36(8):1532--1545, 2014.

\bibitem{drucker1992improving}
Harris Drucker and Yann Le~Cun.
\newblock {I}mproving {G}eneralization {P}erformance using {D}ouble
  {B}ackpropagation.
\newblock {\em IEEE Transactions on Neural Networks}, 3(6):991--997, 1992.

\bibitem{guo2016ms}
Yandong Guo, Lei Zhang, Yuxiao Hu, Xiaodong He, and Jianfeng Gao.
\newblock Ms-celeb-1m: A dataset and benchmark for large-scale face
  recognition.
\newblock In {\em The European Conference on Computer Vision}, pages 87--102,
  2016.

\bibitem{guo2018double}
Yong Guo, Qingyao Wu, Chaorui Deng, Jian Chen, and Mingkui Tan.
\newblock Double forward propagation for memorized batch normalization.
\newblock In {\em AAAI Conference on Artificial Intelligence}, pages
  3134--3141, 2018.

\bibitem{he2015deep}
Kaiming He, Xiangyu Zhang, Shaoqing Ren, and Jian Sun.
\newblock {D}eep {R}esidual {L}earning for {I}mage {R}ecognition.
\newblock In {\em The IEEE Conference on Computer Vision and Pattern
  Recognition}, pages 770--778, 2016.

\bibitem{he2016identity}
Kaiming He, Xiangyu Zhang, Shaoqing Ren, and Jian Sun.
\newblock {I}dentity {M}appings in {D}eep {R}esidual {N}etworks.
\newblock In {\em The European Conference on Computer Vision}, pages 630--645,
  2016.

\bibitem{he2018progressive}
Yang He, Xuanyi Dong, Guoliang Kang, Yanwei Fu, and Yi~Yang.
\newblock Progressive deep neural networks acceleration via soft filter
  pruning.
\newblock {\em arXiv preprint}, abs/1808.07471, 2018.

\bibitem{he2018soft}
Yang He, Guoliang Kang, Xuanyi Dong, Yanwei Fu, and Yi~Yang.
\newblock Soft filter pruning for accelerating deep convolutional neural
  networks.
\newblock In {\em Proceedings of the International Joint Conference on
  Artificial Intelligence}, pages 2234--2240, 2018.

\bibitem{he2019filter}
Yang He, Ping Liu, Ziwei Wang, Zhilan Hu, and Yi~Yang.
\newblock Filter pruning via geometric median for deep convolutional neural
  networks acceleration.
\newblock In {\em The IEEE Conference on Computer Vision and Pattern
  Recognition}, pages 4340--4349, 2019.

\bibitem{he2017channel}
Yihui He, Xiangyu Zhang, and Jian Sun.
\newblock {C}hannel {P}runing for {A}ccelerating {V}ery {D}eep {N}eural
  {N}etworks.
\newblock In {\em The IEEE International Conference on Computer Vision}, pages
  1398--1406, 2017.

\bibitem{howard2019searching}
Andrew Howard, Mark Sandler, Grace Chu, Liang-Chieh Chen, Bo~Chen, Mingxing
  Tan, Weijun Wang, Yukun Zhu, Ruoming Pang, Vijay Vasudevan, et~al.
\newblock Searching for mobilenetv3.
\newblock {\em arXiv preprint}, abs/1905.02244, 2019.

\bibitem{howard2017mobilenets}
Andrew~G Howard, Menglong Zhu, Bo~Chen, Dmitry Kalenichenko, Weijun Wang,
  Tobias Weyand, Marco Andreetto, and Hartwig Adam.
\newblock Mobilenets: Efficient convolutional neural networks for mobile vision
  applications.
\newblock {\em arXiv preprint}, abs/1704.04861, 2017.

\bibitem{huang2016densely}
Gao Huang, Zhuang Liu, Laurens Van Der~Maaten, and Kilian~Q Weinberger.
\newblock {D}ensely {C}onnected {C}onvolutional {N}etworks.
\newblock In {\em The IEEE Conference on Computer Vision and Pattern
  Recognition}, pages 4700--4708, 2017.

\bibitem{DBLP:journals/corr/HuangSLSW16}
Gao Huang, Yu~Sun, Zhuang Liu, Daniel Sedra, and Kilian~Q. Weinberger.
\newblock {D}eep {N}etworks with {S}tochastic {D}epth.
\newblock In {\em The European Conference on Computer Vision}, pages 646--661,
  2016.

\bibitem{huang2008labeled}
Gary~B Huang, Marwan Mattar, Tamara Berg, and Eric Learned-Miller.
\newblock Labeled faces in the wild: A database forstudying face recognition in
  unconstrained environments.
\newblock 2008.

\bibitem{kemelmacher2016megaface}
Ira Kemelmacher-Shlizerman, Steven~M Seitz, Daniel Miller, and Evan Brossard.
\newblock The megaface benchmark: 1 million faces for recognition at scale.
\newblock In {\em The IEEE Conference on Computer Vision and Pattern
  Recognition}, pages 4873--4882, 2016.

\bibitem{krizhevsky2009learning}
Alex Krizhevsky and Geoffrey Hinton.
\newblock {L}earning {M}ultiple {L}ayers of {F}eatures from {T}iny {I}mages,
  2009.

\bibitem{krizhevsky2012imagenet}
Alex Krizhevsky, Ilya Sutskever, and Geoffrey~E Hinton.
\newblock {I}magenet {C}lassification with {D}eep {C}onvolutional {N}eural
  {N}etworks.
\newblock In {\em Advances in Neural Information Processing Systems}, pages
  1097--1105, 2012.

\bibitem{lecun2015deep}
Yann LeCun, Yoshua Bengio, and Geoffrey Hinton.
\newblock {D}eep {L}earning.
\newblock {\em Nature}, 521(7553):436--444, 2015.

\bibitem{lecun1989backpropagation}
Yann LeCun, Bernhard Boser, John~S Denker, Donnie Henderson, Richard~E Howard,
  Wayne Hubbard, and Lawrence~D Jackel.
\newblock {B}ackpropagation {A}pplied to {H}andwritten {z}ip {C}ode
  {R}ecognition.
\newblock {\em Neural Computation}, 1(4):541--551, 1989.

\bibitem{Lee2015}
Chen-Yu Lee, Saining Xie, Patrick Gallagher, Zhengyou Zhang, and Zhuowen Tu.
\newblock {D}eeply-supervised {N}ets.
\newblock In {\em International Conference on Artificial Intelligence and
  Statistics}, 2015.

\bibitem{li2016pruning}
Hao Li, Asim Kadav, Igor Durdanovic, Hanan Samet, and Hans~Peter Graf.
\newblock {P}runing {F}ilters for {E}fficient {C}onvnets.
\newblock In {\em International Conference on Learning Representations}, 2017.

\bibitem{liu2017sphereface}
Weiyang Liu, Yandong Wen, Zhiding Yu, Ming Li, Bhiksha Raj, and Le~Song.
\newblock Sphereface: Deep hypersphere embedding for face recognition.
\newblock In {\em The IEEE Conference on Computer Vision and Pattern
  Recognition}, pages 212--220, 2017.

\bibitem{luo2017thinet}
Jian-Hao Luo, Jianxin Wu, and Weiyao Lin.
\newblock {T}hinet: {A} {F}ilter {L}evel {P}runing {M}ethod for {D}eep {N}eural
  {N}etwork {C}ompression.
\newblock In {\em The IEEE International Conference on Computer Vision}, pages
  5068--5076, 2017.

\bibitem{molchanov2019importance}
Pavlo Molchanov, Arun Mallya, Stephen Tyree, Iuri Frosio, and Jan Kautz.
\newblock Importance estimation for neural network pruning.
\newblock In {\em The IEEE Conference on Computer Vision and Pattern
  Recognition}, pages 11264--11272, 2019.

\bibitem{moschoglou2017agedb}
Stylianos Moschoglou, Athanasios Papaioannou, Christos Sagonas, Jiankang Deng,
  Irene Kotsia, and Stefanos Zafeiriou.
\newblock Agedb: the first manually collected, in-the-wild age database.
\newblock In {\em The IEEE Conference on Computer Vision and Pattern
  Recognition Workshops}, pages 51--59, 2017.

\bibitem{nair2010rectified}
Vinod Nair and Geoffrey~E Hinton.
\newblock {R}ectified {L}inear {U}nits {I}mprove {R}estricted {B}oltzmann
  {M}achines.
\newblock In {\em International Conference on Machine Learning}, pages
  807--814, 2010.

\bibitem{DBLP:journals/pami/RanjanPC19}
Rajeev Ranjan, Vishal~M. Patel, and Rama Chellappa.
\newblock Hyperface: {A} deep multi-task learning framework for face detection,
  landmark localization, pose estimation, and gender recognition.
\newblock {\em IEEE Transactions on Pattern Analysis and Machine Intelligence},
  41(1):121--135, 2019.

\bibitem{DBLP:journals/pami/RenHG017}
Shaoqing Ren, Kaiming He, Ross~B. Girshick, and Jian Sun.
\newblock Faster {R-CNN:} towards real-time object detection with region
  proposal networks.
\newblock {\em IEEE Transactions on Pattern Analysis and Machine Intelligence},
  39(6):1137--1149, 2017.

\bibitem{russakovsky2015imagenet}
Olga Russakovsky, Jia Deng, Hao Su, Jonathan Krause, Sanjeev Satheesh, Sean Ma,
  Zhiheng Huang, Andrej Karpathy, Aditya Khosla, Michael Bernstein, et~al.
\newblock {I}magenet {L}arge {S}cale {V}isual {R}ecognition {C}hallenge.
\newblock {\em International Journal of Computer Vision}, 115(3):211--252,
  2015.

\bibitem{sandler2018mobilenetv2}
Mark Sandler, Andrew Howard, Menglong Zhu, Andrey Zhmoginov, and Liang-Chieh
  Chen.
\newblock Mobilenetv2: Inverted residuals and linear bottlenecks.
\newblock In {\em The IEEE Conference on Computer Vision and Pattern
  Recognition}, pages 4510--4520, 2018.

\bibitem{schroff2015facenet}
Florian Schroff, Dmitry Kalenichenko, and James Philbin.
\newblock {F}acenet: {A} {U}nified {E}mbedding for {F}ace {R}ecognition and
  {C}lustering.
\newblock In {\em The IEEE Conference on Computer Vision and Pattern
  Recognition}, pages 815--823, 2015.

\bibitem{sengupta2016frontal}
Soumyadip Sengupta, Jun-Cheng Chen, Carlos Castillo, Vishal~M Patel, Rama
  Chellappa, and David~W Jacobs.
\newblock Frontal to profile face verification in the wild.
\newblock In {\em IEEE Winter Conference on Applications of Computer Vision},
  pages 1--9, 2016.

\bibitem{DBLP:journals/pami/ShelhamerLD17}
Evan Shelhamer, Jonathan Long, and Trevor Darrell.
\newblock Fully convolutional networks for semantic segmentation.
\newblock {\em IEEE Transactions on Pattern Analysis and Machine Intelligence},
  39(4):640--651, 2017.

\bibitem{shen2015learning}
Li~Shen, Zhouchen Lin, and Qingming Huang.
\newblock {R}elay {B}ackpropagation for {E}ffective {L}earning of {D}eep
  {C}onvolutional {N}eural {N}etworks.
\newblock In {\em The European Conference on Computer Vision}, pages 467--482,
  2016.

\bibitem{simonyan2014very}
Karen Simonyan and Andrew Zisserman.
\newblock {V}ery {D}eep {C}onvolutional {N}etworks for {L}arge-scale {I}mage
  {R}ecognition.
\newblock In {\em International Conference on Learning Representations}, 2015.

\bibitem{srivastava2015training}
Rupesh~K Srivastava, Klaus Greff, and J{\"u}rgen Schmidhuber.
\newblock {T}raining {V}ery {D}eep {N}etworks.
\newblock In {\em Advances in Neural Information Processing Systems}, pages
  2377--2385, 2015.

\bibitem{srivastava2015highway}
Rupesh~Kumar Srivastava, Klaus Greff, and J{\"u}rgen Schmidhuber.
\newblock {H}ighway {N}etworks.
\newblock In {\em International Conference on Machine Learning Deep Learning
  Workshop}, 2015.

\bibitem{sun2015deeply}
Yi~Sun, Xiaogang Wang, and Xiaoou Tang.
\newblock {D}eeply {L}earned {F}ace {R}epresentations are {S}parse,
  {S}elective, and {R}obust.
\newblock In {\em The IEEE Conference on Computer Vision and Pattern
  Recognition}, pages 2892--2900, 2015.

\bibitem{DBLP:journals/corr/SzegedyIV16}
Christian Szegedy, Sergey Ioffe, Vincent Vanhoucke, and Alexander~A. Alemi.
\newblock {I}nception-v4, {I}nception-{R}esnet and the {I}mpact of {R}esidual
  {C}onnections on learning.
\newblock In {\em {AAAI} Conference on Artificial Intelligence}, pages
  4278--4284, 2017.

\bibitem{szegedy2015going}
Christian Szegedy, Wei Liu, Yangqing Jia, Pierre Sermanet, Scott Reed, Dragomir
  Anguelov, Dumitru Erhan, Vincent Vanhoucke, and Andrew Rabinovich.
\newblock {G}oing {D}eeper with {C}onvolutions.
\newblock In {\em The IEEE Conference on Computer Vision and Pattern
  Recognition}, pages 1--9, 2015.

\bibitem{tan2019mnasnet}
Mingxing Tan, Bo~Chen, Ruoming Pang, Vijay Vasudevan, Mark Sandler, Andrew
  Howard, and Quoc~V Le.
\newblock Mnasnet: Platform-aware neural architecture search for mobile.
\newblock In {\em The IEEE Conference on Computer Vision and Pattern
  Recognition}, pages 2820--2828, 2019.

\bibitem{teerapittayanon2016branchynet}
Surat Teerapittayanon, Bradley McDanel, and H~Kung.
\newblock {B}ranchynet: {F}ast {I}nference via {E}arly {E}xiting from {D}eep
  {N}eural {N}etworks.
\newblock In {\em International Conference on Pattern Recognition}, 2016.

\bibitem{wang2018cosface}
Hao Wang, Yitong Wang, Zheng Zhou, Xing Ji, Dihong Gong, Jingchao Zhou, Zhifeng
  Li, and Wei Liu.
\newblock Cosface: Large margin cosine loss for deep face recognition.
\newblock In {\em The IEEE Conference on Computer Vision and Pattern
  Recognition}, pages 5265--5274, 2018.

\bibitem{DBLP:journals/pami/WangWSDH18}
Peng Wang, Qi~Wu, Chunhua Shen, Anthony~R. Dick, and Anton van~den Hengel.
\newblock {FVQA:} fact-based visual question answering.
\newblock {\em IEEE Transactions on Pattern Analysis and Machine Intelligence},
  40(10):2413--2427, 2018.

\bibitem{DBLP:journals/pami/WeiXLHNDZY16}
Yunchao Wei, Wei Xia, Min Lin, Junshi Huang, Bingbing Ni, Jian Dong, Yao Zhao,
  and Shuicheng Yan.
\newblock {HCP:} {A} flexible {CNN} framework for multi-label image
  classification.
\newblock {\em IEEE Transactions on Pattern Analysis and Machine Intelligence},
  38(9):1901--1907, 2016.

\bibitem{Wilson2003The}
D.~R. Wilson and T.~R. Martinez.
\newblock {T}he {G}eneral {I}nefficiency of {B}atch {T}raining for {G}radient
  {D}escent {L}earning.
\newblock {\em Neural networks}, 16(10):1429, 2003.

\bibitem{xie2017aggregated}
Saining Xie, Ross Girshick, Piotr Doll{\'a}r, Zhuowen Tu, and Kaiming He.
\newblock Aggregated residual transformations for deep neural networks.
\newblock In {\em The IEEE Conference on Computer Vision and Pattern
  Recognition}, pages 1492--1500, 2017.

\bibitem{ye2018rethinking}
Jianbo Ye, Xin Lu, Zhe Lin, and James~Z Wang.
\newblock Rethinking the smaller-norm-less-informative assumption in channel
  pruning of convolution layers.
\newblock In {\em International Conference on Learning Representations}, 2018.

\bibitem{yu2018nisp}
Ruichi Yu, Ang Li, Chun-Fu Chen, Jui-Hsin Lai, Vlad~I Morariu, Xintong Han,
  Mingfei Gao, Ching-Yung Lin, and Larry~S Davis.
\newblock Nisp: Pruning networks using neuron importance score propagation.
\newblock In {\em The IEEE Conference on Computer Vision and Pattern
  Recognition}, pages 9194--9203, 2018.

\bibitem{zagoruyko2016wide}
Sergey Zagoruyko and Nikos Komodakis.
\newblock {W}ide {R}esidual {N}etworks.
\newblock In {\em British Machine Vision Conference}, 2016.

\bibitem{zhang2018shufflenet}
Xiangyu Zhang, Xinyu Zhou, Mengxiao Lin, and Jian Sun.
\newblock Shufflenet: An extremely efficient convolutional neural network for
  mobile devices.
\newblock In {\em The IEEE Conference on Computer Vision and Pattern
  Recognition}, pages 6848--6856, 2018.

\bibitem{zhuang2018discrimination}
Zhuangwei Zhuang, Mingkui Tan, Bohan Zhuang, Jing Liu, Yong Guo, Qingyao Wu,
  Junzhou Huang, and Jinhui Zhu.
\newblock Discrimination-aware channel pruning for deep neural networks.
\newblock In {\em Advances in Neural Information Processing Systems}, pages
  875--886, 2018.

\end{thebibliography}
}

\end{document}